\documentclass[runningheads]{llncs}

\usepackage{eccv}
\usepackage{eccvabbrv}

\usepackage{graphicx}
\usepackage{lmodern}
\usepackage{booktabs}
\usepackage{algorithm}
\usepackage{algpseudocode}
\usepackage{multirow}
\usepackage{wrapfig}
\usepackage{pifont}
\usepackage[accsupp]{axessibility}

\usepackage{hyperref}
\usepackage{orcidlink}

\newcommand{\bwm}{A2World}
\newcommand{\arwm}{A2World-sim}
\newcommand{\vlva}{A2World-policy}

\newcommand{\equalcontrib}{\textsuperscript{*}}
\newcommand{\corrauth}{\textsuperscript{\dag}}

\begin{document}

\title{Learning Transferable Dynamics Priors from Action to World Modeling}

\titlerunning{Learning Transferable Dynamics Priors from Action to World Modeling}

\author{
Ze Huang\inst{1}\equalcontrib \and
Jiahui Zhang\inst{1}\equalcontrib \and
Hairuo Liu\inst{2,3}\equalcontrib \and
Chenxi Zhang\inst{2} \and
Ran Cheng\inst{4} \and
Li Zhang\inst{1,2}\corrauth
}

\authorrunning{Huang et al.}

\institute{
School of Data Science, Fudan University, Shanghai, China
\and
Shanghai Innovation Institute, Shanghai, China
\and
Shanghai Jiao Tong University, Shanghai, China
\and
McGill University, Montreal, QC, Canada
}

\maketitle

\begingroup
\renewcommand\thefootnote{}
\footnotetext{
\equalcontrib\ Equal contribution.
\corrauth\ Corresponding author:
\href{mailto:lizhangfd@fudan.edu.cn}{lizhangfd@fudan.edu.cn}
}
\endgroup

\vspace{-1.2em}
\begin{center}
\small
\href{https://github.com/LogosRoboticsGroup/A2World}
{\texttt{https://github.com/LogosRoboticsGroup/A2World}}
\end{center}

\includegraphics[width=1.0\linewidth]{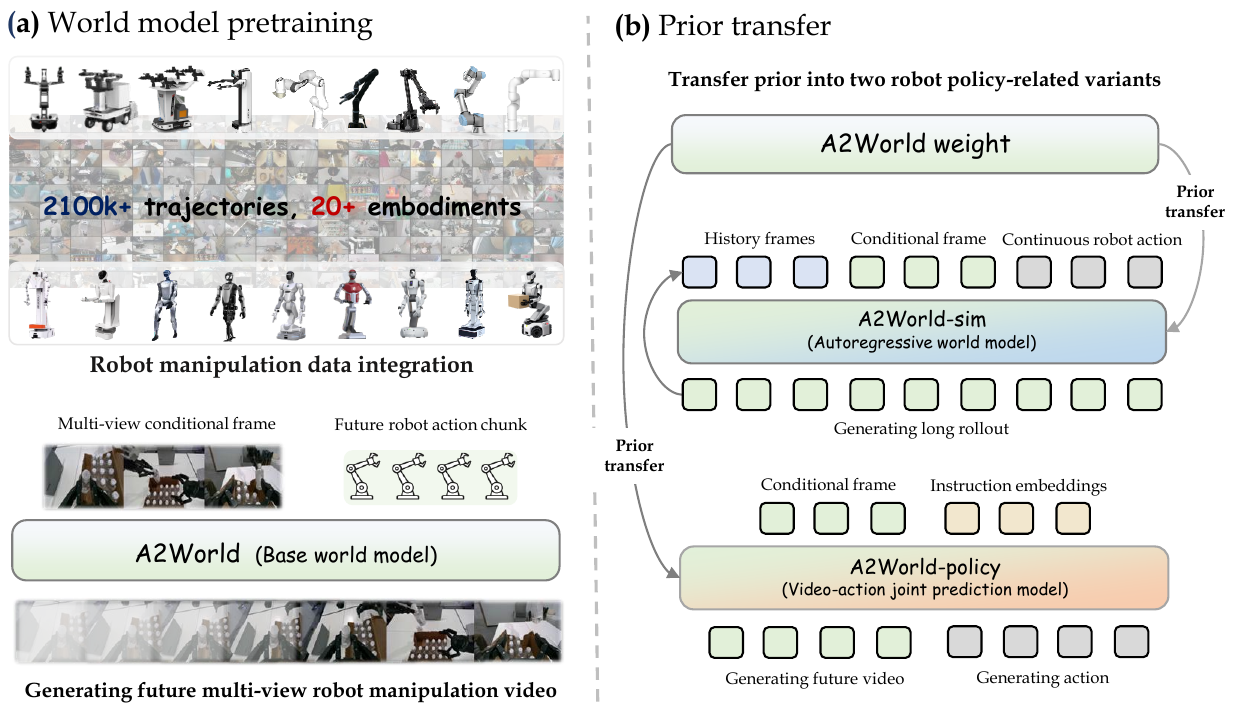} 
\captionof{figure}{
\textbf{We view action-conditioned world modeling as a transferable dynamics prior for robot learning.} (a) We curate 2,100k+ robot manipulation trajectories spanning 20+ embodiments, and pretrain a DiT-based multi-view base world model (\bwm{}) to predict future spatiotemporally consistent manipulation videos from an initial frame and a future action chunk. (b) The pretrained weights are then adapted as a reusable prior into two policy-related variants: \arwm{}, a long-horizon autoregressive simulator for policy evaluation, and \vlva{}, a video-action joint prediction model instantiated as an instruction-conditioned robot policy.}
\label{fig:teaser} 

\begin{abstract}
We study action-conditioned world modeling as a scalable way to learn transferable dynamics priors for robot learning.
By pretraining a model to predict how actions drive visual scene evolution, the resulting world model captures reusable interaction dynamics beyond appearance-level video generation.
Concretely, we pretrain a multi-view interactive base diffusion world model, \bwm{}, on large-scale robot manipulation data with real action annotations.
We validate the learned dynamics priors from two complementary perspectives.
First, we adapt \bwm{} into a task- or scene-specialized real-world simulator, \arwm{}, whose long-horizon rollouts support simulator-based policy evaluation and scalable what-if analysis by replacing real-robot rollouts with world model rollouts.
Second, starting from the same pretrained weights, we adapt \bwm{} into a video-action joint prediction model, \vlva{}, that predicts actions under visual and instruction conditioning.
Experiments across simulation benchmarks and real-robot settings demonstrate that action-conditioned world model pretraining yields transferable dynamics priors that benefit both simulator-centric and policy-centric robot learning.
\end{abstract}
\section{Introduction}
\label{sec:intro}
Recent robot learning methods increasingly build on video generative models~\cite{agarwal2025cosmos, wan2025wan, hacohen2024ltx, blattmann2023stable}, and existing efforts have largely evolved along two directions.
One line of work adapts these models into vision-language-action (VLA) policies~\cite{liao2025genie,hu2024video,wen2024vidman,he2024learning, luo2024grounding,wu2023unleashing}, with recent advances further moving toward joint video-action prediction~\cite{bi2025motus, kim2026cosmos, dreamzero2025, lingbot-va2026, zhu2025unified, pai2025mimicvideo, li2025unified}.
The other line develops action-conditioned world models for data augmentation and policy evaluation~\cite{guo2025ctrl, zhu2025irasim, gao2026dreamdojo, jiang2025enerverse, team2025evaluating, jang2025dreamgen}, and more recently combines them with reward models and reinforcement learning for policy post-training~\cite{zhang2025reinforcing, jiang2025world4rl, zhu2025wmpo, li2025vla}.

Despite this progress, current approaches still underexploit robot-data pretraining as a source of transferable dynamics priors.
Many works~\cite{kim2026cosmos, guo2025ctrl, zhu2025irasim, jiang2025world4rl} directly fine-tune from generic video-generation checkpoints, without large-scale pretraining on action-grounded robot data.
Others~\cite{liao2025genie, gao2026dreamdojo} pretrain on a single dataset, limiting the embodiments, camera setups, and motion patterns the model can absorb.
Even large-scale efforts~\cite{bi2025motus, lingbot-va2026, zhang2025reinforcing} are often optimized for a single downstream objective, rather than explicitly designed for transferable reuse across simulator-centric and policy-centric pipelines.
As a result, the field has yet to establish a unified robot-data pretraining paradigm that learns reusable action-to-dynamics priors and reliably benefits both world model-based simulation and downstream policy learning.

In this paper, we study action-conditioned world modeling as a scalable route to learn transferable dynamics priors for robotics.
The key intuition is that actions provide a natural causal supervision signal in manipulation: while objects, scenes, and viewpoints vary widely across datasets, the underlying interaction rules are governed by how actions induce state changes (e.g., contacts, grasps, pushes, and releases). 
Pretraining a model to predict visual scene evolution conditioned on actions therefore forces it to encode controllable, action-grounded dynamics beyond appearance-level video prediction, yielding reusable interaction priors. 
Leveraging the recent surge of openly available high-quality robot manipulation datasets~\cite{bu2025agibot, tian2025interndata, wu2025robocoin, jiang2025galaxea, contributors2025internroboticsrepo}, we pretrain a multi-view interactive diffusion world model, \bwm{}, directly on real robot action annotations. 
We do not rely on auxiliary latent-action models~\cite{bi2025motus, dreamzero2025, gao2026dreamdojo} to produce indirect pseudo labels, encouraging generalization across embodiments, camera setups, tasks, and motion patterns.
We validate the value of this pretraining from two complementary perspectives that are central to robot learning.
First, on the simulation side, we fine-tune \bwm{} into a history-aware autoregressive world model, \arwm{}, and specialize it as a task- or scene-specific real-world simulator.
\arwm{} supports long-horizon rollouts for simulator-based robot policy evaluation by replacing real-world rollouts with world model rollouts, reducing reliance on real-robot interaction during adaptation and assessment.
Second, on the policy side, starting from the same pretrained weights, we transfer \bwm{} into \vlva{}, a MoE-like video-action policy that shares attention across video and action tokens while keeping action-specific denoising branches, enabling action generation to reuse the pretrained visual dynamics prior while retaining dedicated modeling capacity.

The main contributions of this paper are as follows:
\textbf{(i)} We introduce action-to-video world model pretraining as a robot-data-driven approach for learning transferable dynamics priors, and instantiate it with a multi-view interactive diffusion world model, \bwm{}, pretrained on large-scale, high-quality manipulation data spanning diverse embodiments, tasks, environments, and object categories;
\textbf{(ii)} We demonstrate that the learned dynamics prior can be reused for simulator-centric robot learning by adapting \bwm{} into a history-aware autoregressive simulator, \arwm{}, enabling long-horizon action-conditioned rollouts for real-world simulation, simulator-based policy evaluation, and simulator-based post-training;
\textbf{(iii)} We demonstrate that the same dynamics prior can also be reused for policy-centric robot learning by adapting \bwm{} into a MoE-like video-action joint prediction model, \vlva{}, yielding a strong robot policy under visual and instruction conditioning;
\textbf{(iv)} Across simulation benchmarks and our real-robot platform, we show that action-to-video pretraining produces markedly stronger transferable priors than text-conditioned video pretraining and other robot pretraining baselines, translating into consistent downstream gains and policy improvements across tasks.

\section{Related works}
\label{sec:related}

\noindent \textbf{Action-conditioned robot world models}
Action-conditioned robot world models~\cite{guo2025ctrl, zhu2025irasim, jang2025dreamgen, li2025vla, zhu2025wmpo, xiao2025world, gao2026dreamdojo, liao2025genie, wu2024ivideogpt, jiang2025world4rl, team2025evaluating, quevedo2025worldgym, jiang2025enerverse, zhang2025reinforcing} condition on low-level robot signals, such as end-effector pose changes or joint-angle deltas, to generate future manipulation videos.
Early applications of such models focused on data augmentation for improving VLA training~\cite{guo2025ctrl}, or on serving as safe and scalable validators for VLA policies without requiring real-robot execution~\cite{team2025evaluating, liao2025genie}.
More recent work has also started to improve their generalization ability across unseen tasks and environments~\cite{gao2026dreamdojo}.
A subset of world model-related works further treats the learned world model as a simulator, and combines it with reward models and reinforcement learning algorithms for VLA policy post-training~\cite{li2025vla, xiao2025world, zhu2025wmpo, zhang2025reinforcing, jiang2025world4rl}.
Recent efforts have moved beyond validation-only settings (e.g., simply replacing the simulator~\cite{liu2023libero, mees2022calvin} in standard benchmarks with a learned world model), and toward more practical pipelines that couple real-world world model simulation with policy improvement~\cite{zhang2025reinforcing, zhu2025wmpo}.

\noindent \textbf{World-action models}
Instead of predicting control commands directly from a single observation, world-action models~\cite{pai2025mimicvideo,kim2026cosmos,lingbot-va2026,dreamzero2025,li2025unified,hu2024video,liao2025genie, yuan2026fast} predict how the scene will evolve and then convert those predicted visual changes into executable actions.
Cosmos Policy~\cite{kim2026cosmos} illustrates how video models can be adapted for control by fine-tuning them to produce action-relevant predictions.
DreamZero~\cite{dreamzero2025} further reveals the potential of using video models as base models for zero-shot transfer to action generation.
LingBot-VA~\cite{lingbot-va2026} interleaves video and action tokens in an autoregressive diffusion policy and introduces efficient mechanisms for closed-loop control.
In contrast, we start from an action-conditioned video world model and show that its pretrained dynamics prior transfers to a strong policy with only minimal modifications.
This transfer avoids a separate action-only pretraining stage and requires only lightweight policy-specific adaptations beyond joint video–action modeling.

\section{Methodology}
\label{sec:method}

\subsection{\bwm{}: the base robot world model}
\begin{figure*}
    \includegraphics[width=1.0\linewidth]{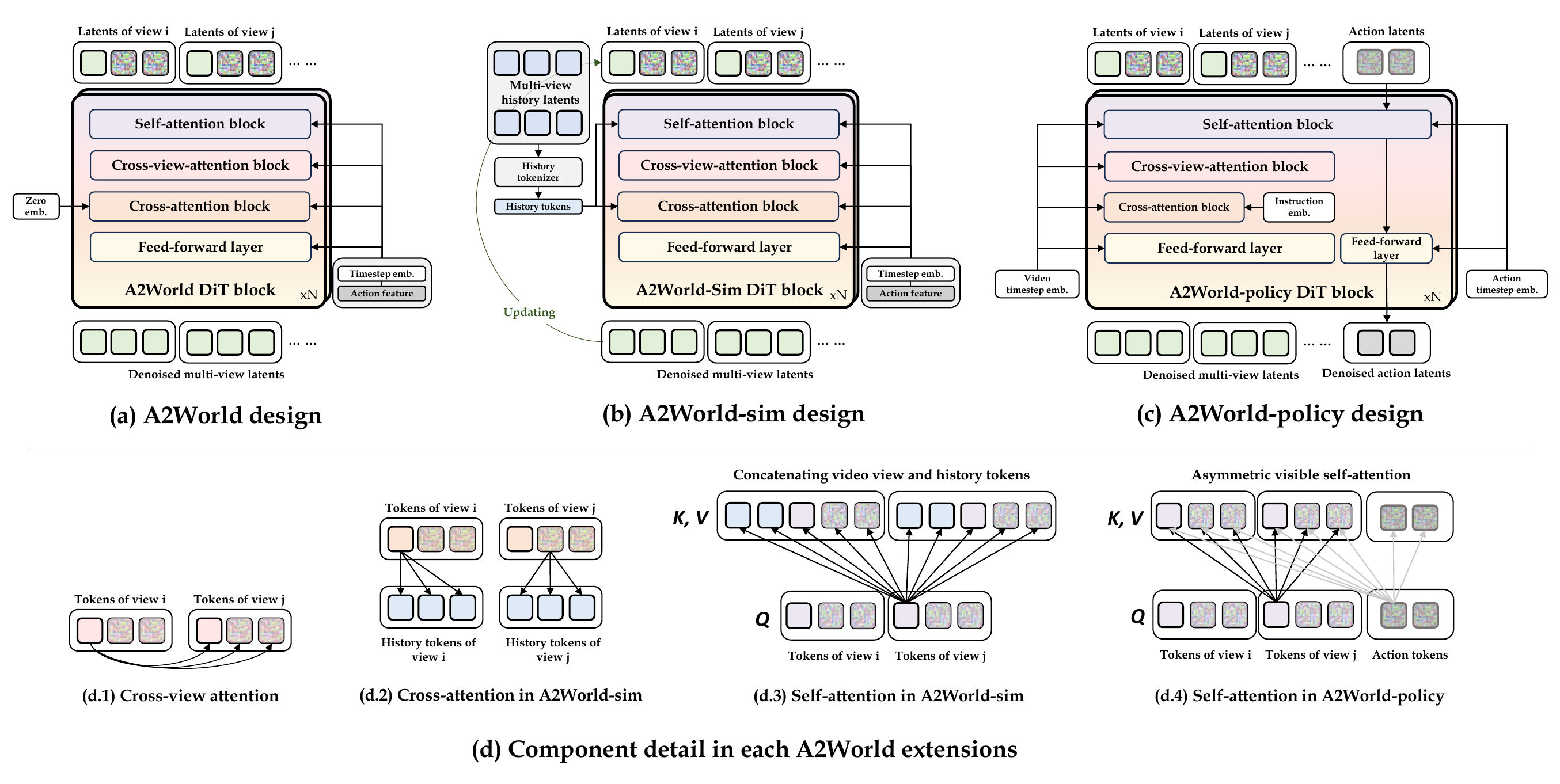}
    \caption{\textbf{Detailed module design of our \bwm{} series.}}
    \label{fig:a2world}
\vspace{-1em}
\end{figure*}

Our base world model (\bwm{}, Fig.~\ref{fig:a2world}.(a)) is designed as an action-conditioned generative world model to learn transferable dynamics priors.
Given a conditioning frame $o_t$ and a chunk of future actions $a_{t+1:t+k}$, it forecasts future observations $o_{t+1:t+k}$:
\begin{equation}
\mathrm{A2World}:p\left(o_{t+1:t+k}\mid o_t, a_{t+1:t+k}\right).
\end{equation}
\noindent\textbf{Action conditioning}
Concretely, the action chunk $a$ is encoded by an MLP, $e=\mathrm{MLP}(a)$, and the resulting action embedding is added to the diffusion timestep embedding used by every DiT block: $\tilde{\tau}(\sigma) = \tau(\sigma) + e$.
$\tau(\sigma)$ denotes the diffusion timestep embedding followed by an MLP projection, and $\tilde{\tau}(\sigma)$ is then consumed by adaptive layer normalization to produce dynamic modulation parameters (scale, shift, and gate).

In this base model, we disable other conditioning inputs by setting the conditional feature to zero, i.e., $\mathbf{c}=\mathbf{0}$, so the model relies solely on action-conditioned temporal modulation.

Specifically, \bwm{} follows a standard latent video diffusion pipeline and operates in the continuous latent space produced by the WAN2.1 tokenizer~\cite{wan2025wan}.
It injects timestep conditions into each DiT block~\cite{peebles2023scalable} via adaptive layer normalization~\cite{ali2025world}.
We train \bwm{} with the EDM denoising score matching objective~\cite{karras2022elucidating} to recover the clean latent sequence from its corrupted version:
\begin{equation}
\mathcal{L}_{\text{A2World}}\left(\sigma\right)=\mathbb{E}_{\mathbf{z}, \mathbf{n}}\left[\|\mathrm{A2World}\left(\mathbf{z}+\mathbf{n} ; \sigma, \mathbf{c}=\mathbf{0}, a\right)-\mathbf{z}\|_{2}^{2}\right],
\end{equation}
where $\mathbf{z}$ is a clean VAE-encoded latent video, $\mathbf{n} \sim \mathcal{N}(0, \sigma ^{2}\mathbf{I})$ is i.i.d. Gaussian noise, and $\sigma$ is the noise level.

\noindent \textbf{Multi-view generation}
Most robot manipulation setups naturally provide multi-view observations, typically including multiple first- or third-view cameras.
Thus, our \bwm{} is implemented to generate multi-view videos jointly.

We denote the multi-view observation as $o_t=\left \{ {o_t^{(v)}} \right \}_{v=1}^{V}$, where $v$ indexes the camera view.
We pack the $V$ views into the temporal dimension and run a single latent video diffusion forward pass:
let $\mathbf{z}_{\text{mv}}$ denote the clean multi-view latent sequence, which we reshape as $\mathbf{z}_{\text{mv}} \in \mathbb{R}^{\mathbf{B}\times \mathbf{C}\times (\mathbf{V}\cdot \mathbf{T})\times \mathbf{H}\times \mathbf{W}}$, so that multi-view generation is processed as temporally concatenated video generation.
To identify different cameras, we introduce learnable view embeddings $\epsilon_{\text{view}}(v) \in \mathbb{R}^{\mathbf{d_{e}}}$ for different views (e.g., first-view vs. third-view), broadcast them over the corresponding spatial-temporal latent grids, and concatenate them to the latent input tokens before patch embedding along the channel dimension:
\begin{equation}
\tilde{\mathbf{z}}_{\text{mv}}^{(v)} = \mathrm{concat}\left(\mathbf{z}^{(v)}_{\text{mv}}, \epsilon_{\text{view}}(v)\right), \quad \tilde{\mathbf{z}}_{\text{mv}}\in \mathbb{R}^{\mathbf{B}\times (\mathbf{C}+\mathbf{d_{e}})\times (\mathbf{V}\cdot \mathbf{T})\times \mathbf{H}\times \mathbf{W}}.
\end{equation}
This provides explicit camera identity to the DiT.
In addition, to enforce cross-view consistency, we insert cross-view attention modules (Fig.~\ref{fig:a2world}.(d.1)) into each DiT block, where tokens from one view attend to tokens from the other views:
\begin{equation}
\tilde{\mathbf{z}}_{\text{mv}}^{(w)} \leftarrow \tilde{\mathbf{z}}_{\text{mv}}^{(w)} + \mathrm{CrossViewAttn}\left( \tilde{\mathbf{z}}_{\text{mv}}^{(w)}, {\tilde{\mathbf{z}}_{\text{mv}}^{(u)}},{u\neq w} \right).
\end{equation}
This module encourages spatiotemporally consistent rollouts across views, while preserving view-specific details.

For notational convenience, we use $\mathbf{z}$ to denote the multi-view latent tokens, and write $\mathbf{z}^v$ when distinguishing them from action latents in the remainder.

\begin{table}[t]
\centering
\caption{\textbf{Curated and preprocessed datasets used for pretraining in \bwm{} and fine-tuning in \arwm{} and \vlva{}.}}
\label{tab:dataset_details}
\scalebox{0.60}{
\begin{tabular}{lcclc}
\toprule
\textbf{Dataset} & \textbf{Trajectories} & \textbf{Embodiment} & \textbf{Data usage}  & \textbf{Used in} \\
\midrule
AgiBot~\cite{bu2025agibot}   & 1003k & Genie-1     & Pretraining  & \bwm{} \\
DROID~\cite{khazatsky2024droid}   & 76k &  Franka   & Pretraining  & \bwm{}\\
OPEN-X~\cite{o2024open}   & 19k &  UR5, WidowX, X-ARM, Jaco2, Franka  & Pretraining  & \bwm{}\\
InternData-A1~\cite{tian2025interndata}   & 630k & Genie-1, ARX LIFT-2, Franka, Agilex Split Aloha   & Pretraining  & \bwm{}\\
InternData-M1-Agilex~\cite{contributors2025internroboticsrepo}   & 105k &  Dual-arm Franka  & Pretraining  & \bwm{}\\
RoboCoin~\cite{wu2025robocoin}  & 246k & 15 types including  AirBot MMK2, Unitree G1edu-u3, etc.   & Pretraining  & \bwm{}\\
Galaxea~\cite{jiang2025galaxea}  & 77k & Galaxea R1 Lite   & Pretraining  & \bwm{}\\
\midrule
LIBERO~\cite{liu2023libero}   & 6k &  Franka  & Fine-tuning & \arwm{}, \vlva{}\\
LIBERO-Plus-Spatial~\cite{fei2025libero}  & 3.7k & Franka   & Evaluation &  \arwm{}, \vlva{}\\
RoboNet~\cite{dasari2019robonet}  & 100k & Sawyer, Franka, WidowX, KUKA, Baxter, Fetch  & Fine-tuning & \arwm{}\\
Custom data  & 2k &  Dual-arm Flexiv Rizon 4S with Robotiq-2F-85 gripper  & Fine-tuning & \arwm{}, \vlva{} \\
\bottomrule
\end{tabular}
}
\vspace{-1em}
\end{table}

\noindent \textbf{\bwm{} pretraining}
We pretrain \bwm{} on curated high-quality robot manipulation datasets to learn transferable dynamics priors across diverse tasks and environments.
To enable action-conditioned pretraining across embodiments, we unify all actions into a shared dual-arm format, where each arm is represented by 7 dimensions (end-effector pose and gripper state).
For single-arm robots, the missing arm is zero-padded.
Tab.~\ref{tab:dataset_details} summarizes the datasets used.
Since different datasets often have different camera setups and view conventions, directly mixing them can introduce conflicts for multi-view generation.
To mitigate this, we adopt a dataset-consistent batching strategy, i.e., each mini-batch is sampled from a single dataset only.

\subsection{\arwm{}: adapting \bwm{} into a long-horizon simulator}
\bwm{} is pretrained to learn transferable action-to-dynamics priors from large-scale robot data.
To reuse this prior for long-horizon simulation, we transfer \bwm{} into a history-aware autoregressive world model, \arwm{} (Fig.~\ref{fig:a2world}.(b)), which conditions on past observations and sequentially rolls out future observations under actions:
\begin{equation}
\text{A2World-sim}: p \left(o_{t+1:t+k}\mid o_t,\; a_{t+1:t+k},\; \mathcal{H}_{t-1}\right),
\quad \mathcal{H}_{t-1}=o_{1:t-1}.
\end{equation}
\noindent \textbf{Pose-guided history sampling}
Given a history sequence and robot actions, we sample a compact set of frames that covers the executed motion trajectory. Specifically, we compute a weighted arc-length from relative actions and sample frames uniformly along it (Alg.~\ref{alg:hist_sampling}).
This preserves key states, e.g., turning points, under a fixed history budget.
For clarity, we illustrate the single-arm variant here, and the dual-arm case is analogous and provided in the Appendix.

\noindent \textbf{History injection}
After selecting a compact history subset, we inject history into the DiT in two complementary ways.
First, the sampled history latent sequence is tokenized and mapped into history tokens:
\begin{equation}
\mathbf{H}=\mathrm{Tok}_{\text{hist}}(\mathcal{H}[\mathcal{S}])\in\mathbb{R}^{B\times N_h\times D}.
\end{equation}
These tokens replace the generic cross-attention condition (set to zero in pretraining) and are attended by target video latent tokens (Fig.~\ref{fig:a2world}.(d.2)):
\begin{equation}
    \mathbf{z} \leftarrow \mathbf{z} + \mathrm{CrossAttn}\left(\mathbf{z}, \mathbf{H}\right).
\end{equation}
Second, we additionally inject the same history tokens into the self-attention memory path by projecting them to key and value memories and concatenating them with the current latent self-attention (Fig.~\ref{fig:a2world}.(d.3)).
This provides a global history memory injection, allowing target latent tokens to directly interact with compact historical states inside self-attention.

\begin{algorithm}[t]
\caption{Arc-uniform history sampling}
\label{alg:hist_sampling}
\vspace{-0.3em}
\scriptsize
\begin{algorithmic}[1]
\Require History length $T_h$; relative actions $\{\Delta x_r\}_{r=1}^{T_h-1}$ with $\Delta x_r=[\Delta p_r,\Delta \theta_r]$; budget $m$; weights $w_t,w_r>0$
\Ensure Sampled history indices $\mathcal{S}$
\State Choose anchor indices $r_s$ (earliest valid frame) and $r_e=T_h$ (latest frame)
\State Compute weighted step lengths: $d_r \leftarrow \sqrt{w_t\|\Delta p_r\|_2^2 + w_r\|\Delta \theta_r\|_2^2}$ for $r=1,\dots,T_h-1$
\State Compute cumulative arc-length: $A_{r_s}\leftarrow 0$; $A_r \leftarrow \sum_{q=r_s}^{r-1} d_q$ for $r=r_s+1,\dots,r_e$
\State Initialize $\mathcal{S}\leftarrow \{r_s, r_e\}$ and set $n_{\text{mid}} \leftarrow m-2$
\For{$s=1$ to $n_{\text{mid}}$}
    \State $\bar{A}_s \leftarrow A_{r_s} + \frac{s}{m-1}(A_{r_e}-A_{r_s})$
    \State $\hat{r} \leftarrow \arg\min_{r\in(r_s,r_e)} |A_r-\bar{A}_s|$
    \State $\mathcal{S}\leftarrow \mathcal{S}\cup\{\hat{r}\}$
\EndFor
\State \Return chronological indices in $\mathcal{S}$
\end{algorithmic}
\vspace{-0.4em}
\end{algorithm}

\noindent \textbf{Autoregressive generation}
Given an initial observation $o_t$ and future actions, we roll out $\text{A2World-sim}$ in a chunk-wise autoregressive manner.
At each step, the model predicts a future chunk $o_{t+1:t+k}$ conditioned on the current frame, history memory, and action chunk, and the generated frames are then appended to the history buffer and reused as the condition for the next rollout step.
This converts the short-horizon predictor into a long-horizon action-conditioned simulator.

To improve long-horizon stability, we adopt a Self-forcing-style~\cite{huang2025self} training strategy: during training, the model is periodically conditioned on its own generated frames, rather than always using ground-truth frames.
Unlike teacher-forcing distillation, we do not require a separate teacher model, since under action conditioning and a given initial frame, the future trajectory is largely determined by the underlying dynamics.
Self-forcing therefore directly exposes the model to its own rollout errors and trains it to recover from them.

\subsection{\vlva{}: adapting \bwm{} into a robot policy}
To transfer the pretrained dynamics prior to the policy setting, we adapt \bwm{} into a MoE-like vision-action joint prediction model, \vlva{} (Fig.~\ref{fig:a2world}.(c)), which jointly models future observations and actions in a single generative process.
This adaptation instantiates the world model as an instruction-conditioned robot policy that takes a language instruction $l$ and an initial frame $o_t$ as input:
\begin{equation}
\label{eq:a2world_policy}
\text{A2World-policy}:p\left(o_{t+1:t+k}, a_{t+1:t+k}\mid o_t,l \right).
\end{equation}
We encode the instruction $l$ with a pretrained T5 text encoder~\cite{raffel2020exploring} and use the resulting token embeddings $\mathbf{h}_l$ as the cross-attention context in each DiT block.
We model future observation latents and future action tokens jointly under the same conditional context given the initial frame and instruction, so that the pretrained world model can be transferred to both simulator-oriented and policy-oriented downstream tasks with one initialization.

\noindent \textbf{Joint diffusion formulation}
Let $\mathbf{z}^v$ and $\mathbf{z}^a$ denote clean video and action latents, respectively.
We add Gaussian perturbations independently:
\begin{equation}
\mathbf{z}^v_{\sigma_v}=\mathbf{z}^v+\mathbf{n}^v,\quad \mathbf{n}^v\sim\mathcal{N}(0,\sigma_v^{2}\mathbf{I}),
\qquad
\mathbf{z}^a_{\sigma_a}=\mathbf{z}^a+\mathbf{n}^a,\quad \mathbf{n}^a\sim\mathcal{N}(0,\sigma_a^{2}\mathbf{I}),
\end{equation}
where $\sigma_v$ and $\sigma_a$ are the modality-specific noise levels.
In our shared-timestep training variant, a single base noise level $\sigma_{\mathrm{base}}$ is sampled and then scaled per modality $\sigma_v=\alpha_v\sigma_{\mathrm{base}}$, $\sigma_a=\alpha_a\sigma_{\mathrm{base}}$,
which improves video-action alignment while preserving modality-specific noise scales.

\noindent \textbf{MoE-like video-action blocks}
Let \(\mathbf{z}_v^\ell\) and \(\mathbf{z}_a^\ell\) denote video and action tokens at block \(\ell\).
An \vlva{} block shares one self-attention module across modalities (Fig.~\ref{fig:a2world}.(d.4)), with modality-specific AdaLN and MLP branches.
We refer to this as MoE-like because video and action share the attention expert for interaction, while each modality keeps its own lightweight denoising branch.
Video tokens are updated by each attention layer.
Action tokens attend to both video and action tokens via the shared self-attention.
Both streams then pass through their AdaLN and MLP branches, and the final action sequence is predicted by a linear head on the last-layer action tokens.

\noindent \textbf{Training objective and guided inference}
The model predicts clean video and action latents jointly:
\begin{equation}
(\hat{\mathbf{z}}^v,\hat{\mathbf{z}}^a)
=
\text{A2World-policy}\!\left(\mathbf{z}^v+\mathbf{n}^v,\mathbf{z}^a+\mathbf{n}^a;\sigma_v,\sigma_a,\mathbf{c}=\mathbf{h}_l\right),
\end{equation}
with a weighted joint denoising objective:
\begin{equation}
\mathcal{L}_{\text{A2World-policy}} (\sigma_{v}, \sigma_{a})
=
\mathbb{E}_{\mathbf{z}^v,\mathbf{z}^a,\mathbf{n}^v,\mathbf{n}^a}\!\left[
\mathbf{w}(\sigma_v)\|\hat{\mathbf{z}}^v-\mathbf{z}^v\|_2^2
+ \lambda_a\,\mathbf{w}(\sigma_a)\|\hat{\mathbf{z}}^a-\mathbf{z}^a\|_2^2
\right].
\end{equation}
At inference, we use modality-wise classifier-free guidance:
\begin{equation}
\hat{\mathbf{z}}_{\mathrm{cfg}}^m
=
\hat{\mathbf{z}}_{\mathrm{u}}^m +
s_m\!\left(\hat{\mathbf{z}}_{\mathrm{c}}^m-\hat{\mathbf{z}}_{\mathrm{u}}^m\right),\quad m\in\{v,a\},
\end{equation}
where $s_v$ and $s_a$ can be set separately for video and action, enabling controllable trade-offs between visual fidelity and action accuracy.
When a single guidance scalar $\gamma$ is used, we set $s_v=s_a=1+\gamma$.

\section{Experiments}
\label{sec:exp}

\subsection{Experimental setups}
\noindent \textbf{World model training setups}
Our \bwm{} initializes its base DiT backbone from the Cosmos-Predict2-2B-Video2World checkpoint~\cite{agarwal2025cosmos}, while the additional conditioning and multi-view modules are properly initialized, resulting in a total model size of 2.5B parameters.
\bwm{} conditions on the initial frame and, given a 20-step action chunk, generates the next 20 frames.
We pretrain the \bwm{} on a mixture of processed robot manipulation datasets~\cite{bu2025agibot, khazatsky2024droid, o2024open, tian2025interndata, contributors2025internroboticsrepo, wu2025robocoin, jiang2025galaxea}, with a total of 2156k trajectories (Tab.~\ref{tab:dataset_details}).
For pretraining, we train all model parameters with 64 H200 GPUs, using a batch size of 12 per GPU and gradient accumulation of 4 for 2 epochs.
For fine-tuning \arwm{}, we use 8 H200 GPUs with a batch size of 24 and a task-dependent number of steps.
Both stages use the fused Adam optimizer with a learning rate of 1e-4 and weight decay of 0.1.
For history-aware mechanism, we set history length $T_{h} = 20$, $w_{r}=0.3$, $w_{t}=1.0$.
Other parameter settings are detailed in the Appendix.

\noindent \textbf{\vlva{} training setups}
We initialize the \vlva{} from the pretrained \bwm{}, and initialize the cross-attention from the Cosmos-Predict2-2B checkpoint since \bwm{} pretraining sets the cross-attention conditioning to zero.
We initialize action-specific modules by copying the corresponding video-branch parameters from \bwm{} for stable joint fine-tuning.
The resulting \vlva{} has 3.0B parameters.
We train with 32 H200 GPUs, global batch size 256, and learning rate 1e-4, using 20k steps for LIBERO and real-robot fine-tuning.
For OOD policy evaluation, we fine-tune \vlva{} variants on LIBERO for 24k steps and evaluate on LIBERO-Plus Spatial.

\begin{wrapfigure}{r}{0.52\columnwidth}
\vspace{-18pt}
\centering
\includegraphics[width=0.46\columnwidth]{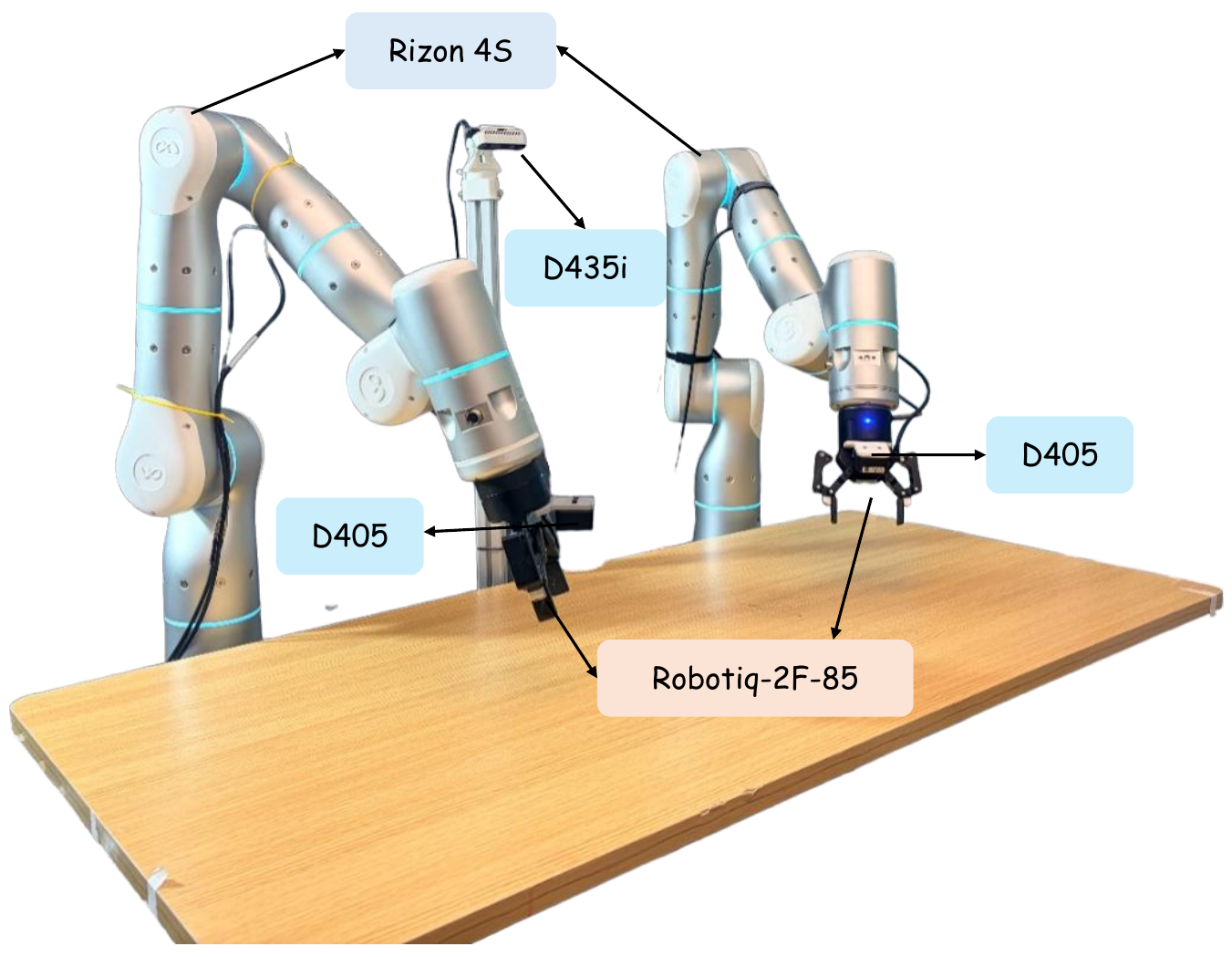}
\caption{\textbf{Our real-robot platform.} Two Flexiv arms with Robotiq-2F-85 grippers are mounted symmetrically at 45° facing the tabletop. The vision system uses one front-view Intel RealSense D435i and two wrist-mounted D405 cameras (480$\times$640, 30fps).}
\label{fig:flexiv}
\vspace{-2pt}
\end{wrapfigure}

\noindent \textbf{Custom real-robot dataset collection}
We constructed a dual-arm manipulation platform based on Flexiv robots, following the setup of Toyota Research Institute (TRI)~\cite{barreiros2025careful}, and conducted data collection through VR teleoperation.
The dataset comprises 5 tasks, i.e., \emph{insert RAM module}, \emph{flip small box}, \emph{toggle power switch}, \emph{lift box high}, \emph{put chain in the box}.
These tasks involve challenging scenarios such as rich-contact manipulation, articulated objects, and deformable objects.
To ensure data quality, we performed careful curation by selecting episodes with minimal idle frames and high task completion rates.
Details are shown in Fig.~\ref{fig:flexiv}.

\subsection{Base world model capability demonstration}
\begin{figure*}
    \includegraphics[width=1.0\linewidth]{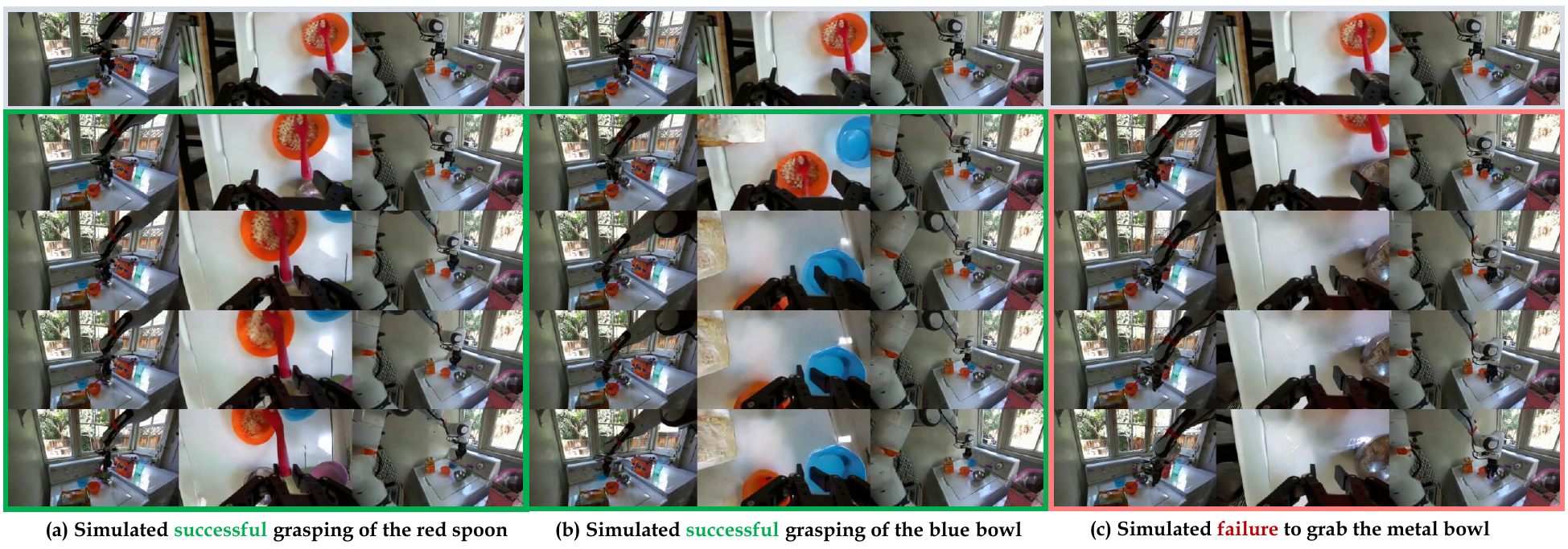}
    \vspace{-2em}
    \caption{\textbf{Rollout presentation on DROID~\cite{khazatsky2024droid} using \bwm{}.} Given an initial frame, \bwm{} can freely simulate grasping different objects in the scene, and can also reproduce a failed grasp attempt under the corresponding action. This suggests that \bwm{} learns to respond to action inputs rather than merely generating `successful' outcomes, which is crucial for downstream counterfactual evaluation.}
    \label{fig:droid_viz}
\vspace{-1em}
\end{figure*}

For our pretrained \bwm{}, we qualitatively demonstrate strong action-conditioned multi-view rollouts.
Fig.~\ref{fig:droid_viz} shows DROID~\cite{khazatsky2024droid} rollouts.
From the same initial observation, \bwm{} can be steered to grasp different objects and can also simulate failures under appropriate actions, suggesting it models action-to-visual dynamics beyond success-only generation.
\begin{figure*}
    \includegraphics[width=1.0\linewidth]{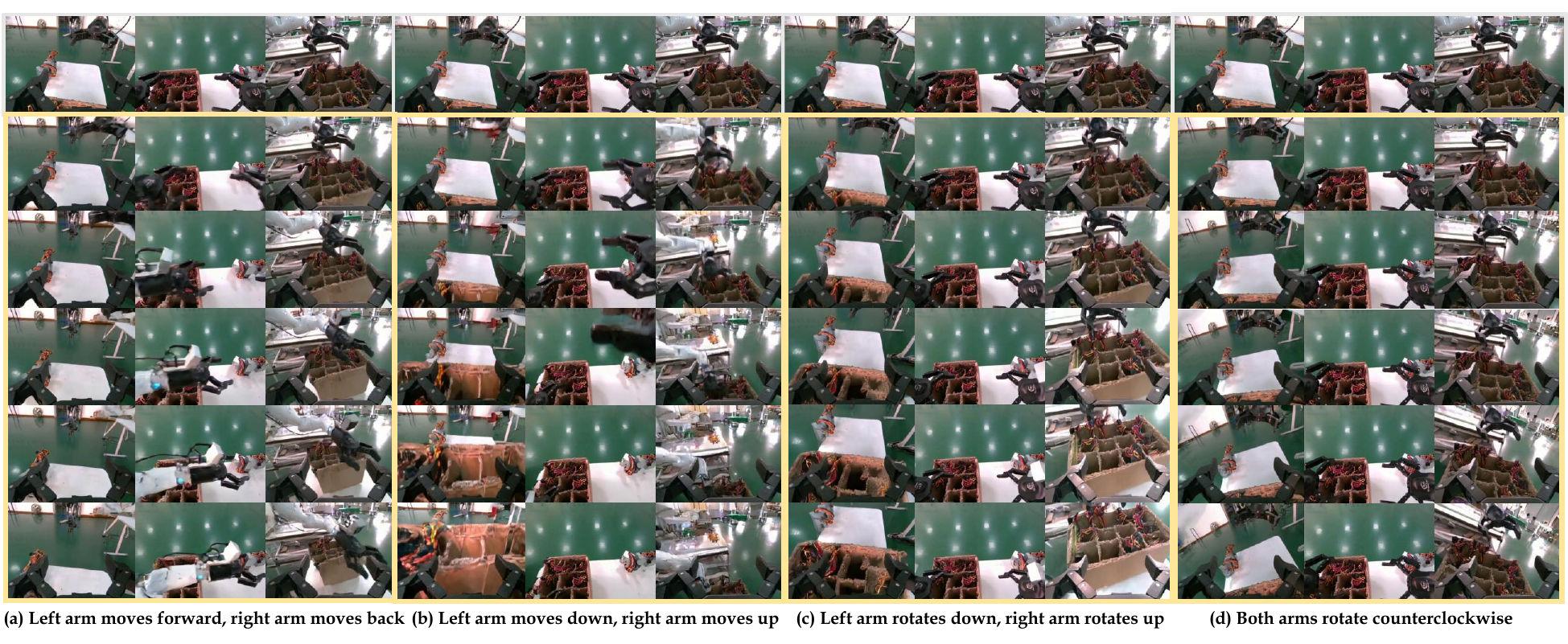}
    \vspace{-2em}
    \caption{\textbf{Full-DoF control of robotic arm on RoboCoin~\cite{wu2025robocoin} using \bwm{}.} Such scripted control never appears in pretraining data, yet \bwm{} can faithfully simulate the resulting scene navigation, indicating strong counterfactual controllability.}
    \label{fig:robocoin_viz}
\vspace{-1em}
\end{figure*}
\begin{figure*}
    \includegraphics[width=1.0\linewidth]{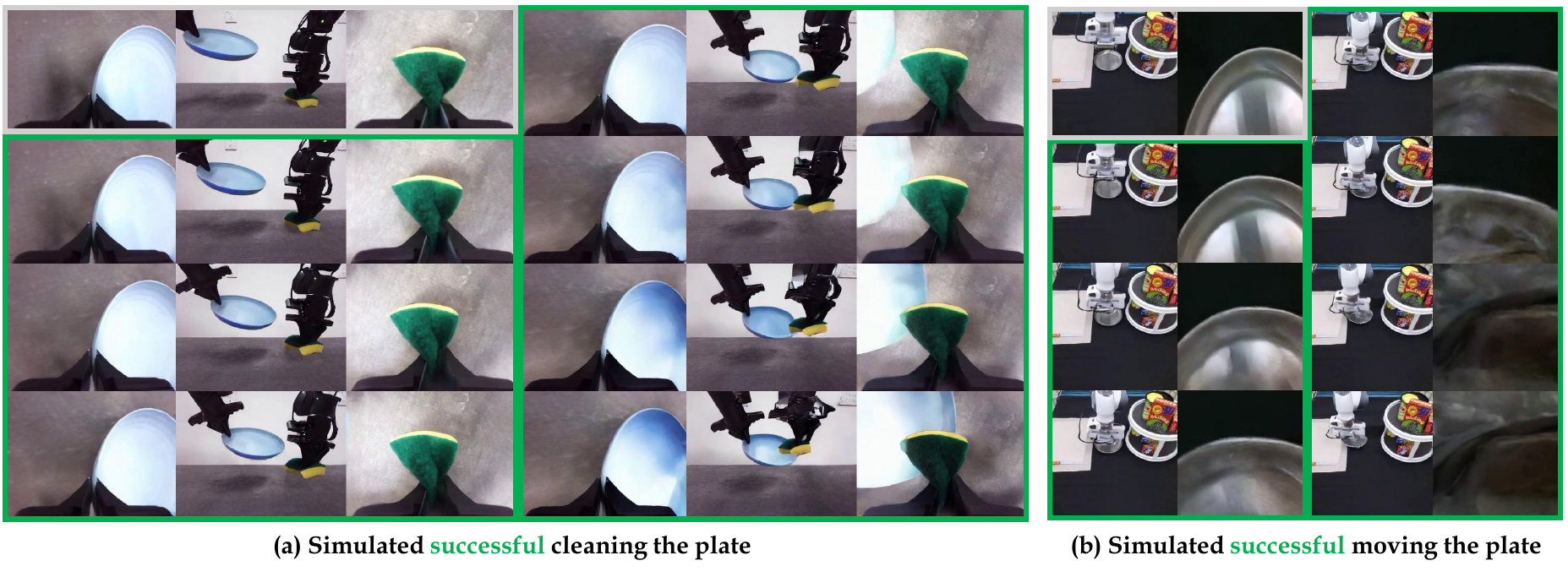}
    \vspace{-2em}
    \caption{\textbf{Out-of-distribution rollouts on RoboMind~\cite{wu2024robomind} (left) and VIOLA~\cite{zhu2023viola} (right) using \bwm{}.} For both RoboMind and VIOLA, the scenes and camera setups are unseen during pretraining.}
    \label{fig:ood_viz}
\vspace{-1em}
\end{figure*}

Fig.~\ref{fig:robocoin_viz} further shows precise full-DoF control on RoboCoin~\cite{wu2025robocoin} with consistent predictions across heterogeneous views.
Finally, Fig.~\ref{fig:ood_viz} presents out-of-distribution (OOD) rollouts on RoboMind~\cite{wu2024robomind} and VIOLA~\cite{zhu2023viola}, where scenes and camera setups are unseen during pretraining.
\bwm{} remains coherent, indicating robust generalization and transferable dynamics priors that we exploit in downstream experiments.

\begin{figure*}
    \includegraphics[width=1.0\linewidth]{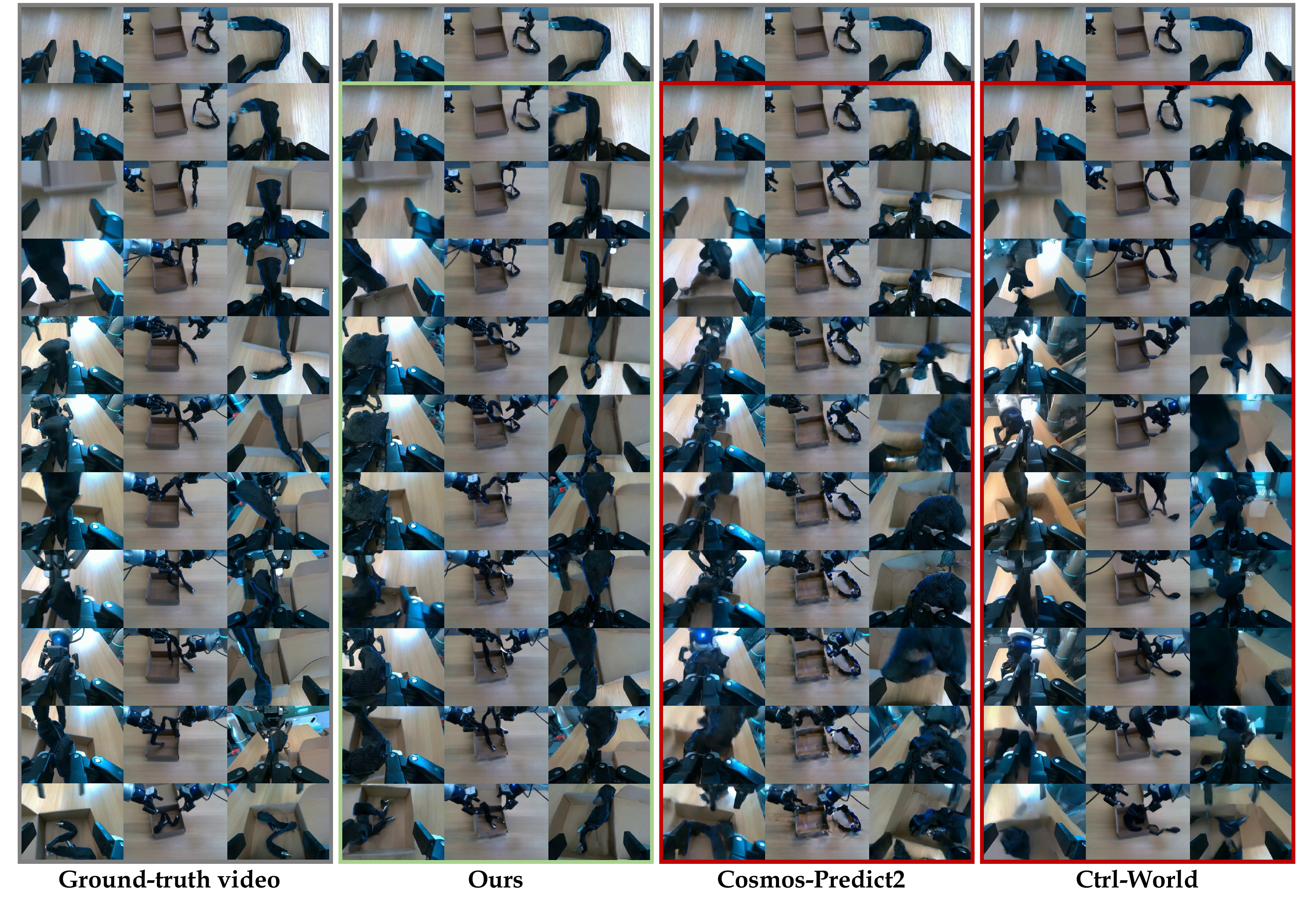}
    \vspace{-2em}
    \caption{\textbf{Qualitative long-rollout generation on \emph{Put chain in the box}.} We autoregressively generate long-horizon videos from same initial frame and real robot actions (frames every 2s). Baselines drift after $\sim$6s and quickly collapse, while our model follows the actions more faithfully, completes the task, and maintains high visual quality.}
    \label{fig:quali_compare}
\vspace{-1.0em}
\end{figure*}

\subsection{Fine-tuned world model evaluation}
\label{sec:arwm-eval}
\noindent \textbf{Rollout quality evaluation}
We fine-tune \bwm{} into \arwm{} on the LIBERO~\cite{liu2023libero} and our custom data for 30k steps, and evaluate generation quality.
We compare against strong pretrained baselines:
\textbf{(i)} Cosmos-Predict2~\cite{agarwal2025cosmos}, using the Cosmos-Predict2-2B-Video2World-480p-10fps model, which is pretrained on broad web-scale data with text conditioning;
\textbf{(ii)} Ctrl-World~\cite{guo2025ctrl}, pretrained on DROID with multi-view modeling;
\textbf{(iii)} Prophet~\cite{zhang2025reinforcing}, which is pretrained with action conditioning but operates in a single-view setting;
\textbf{(iv)} a text-conditioned \bwm{} variant pretrained on the same robot data, which replaces
\begin{table*}[t]
\centering
\small
\setlength{\tabcolsep}{4.5pt}
\caption{\textbf{Rollout quality comparison on simulation and real-robot datasets.}}
\label{tab:wm_eval}
\vspace{-1em}
\scalebox{0.7}{
\begin{tabular}{l|ccccc|ccccc}
\toprule
\multirow{2}{*}{\textbf{Methods}} 
& \multicolumn{5}{c|}{\textbf{LIBERO}~\cite{liu2023libero}} 
& \multicolumn{5}{c}{\textbf{Real Robot (Flexiv Rizon 4S)}} \\
\cmidrule(lr){2-6}\cmidrule(l){7-11}
& PSNR $\uparrow$ & SSIM $\uparrow$ & tSSIM $\uparrow$ &  $\overline{\mathrm{EPE}}$ $\downarrow$ & $\overline{\cos}$  $\uparrow$
& PSNR $\uparrow$ & SSIM $\uparrow$ & tSSIM $\uparrow$ & $\overline{\mathrm{EPE}}$ $\downarrow$ & $\overline{\cos}$ $\uparrow$ \\
\midrule
Cosmos-Predict2~\cite{agarwal2025cosmos} & {25.36} & {.8792} & {.7631} & {.4009} & {.5755} & {24.99} & {.8355} & {.7009} & {.8010} & {.2289} \\
Ctrl-World~\cite{guo2025ctrl} & {23.60} & {.8632} & {.7445} & {.6827} & {.3730} & {21.91} & {.8422} & \textbf{.7158} & {.9689} & {.1195} \\
Prophet~\cite{zhang2025reinforcing} & {26.12} & {.8887} & {.7789} & {.3667} & {.5932} & \underline{25.55} & \underline{.8454} & {.7102} & \underline{.7439} & \underline{.2650} \\
\midrule
\textbf{\arwm{}-T-pre} & \underline{26.18} & \underline{.8892} & \underline{.7794} & \underline{.3533} & \underline{.5973} & {24.64} & {.8198} & {.6411} & {.9328} & {.2293} \\
\textbf{\arwm{}}  & \textbf{26.64} & \textbf{.8957} & \textbf{.7862} & \textbf{.3498} & \textbf{.6045} & \textbf{25.95} & \textbf{.8511} & \underline{.7139} & \textbf{.7218} & \textbf{.2784} \\
\bottomrule
\end{tabular}
}
\vspace{-1em}
\end{table*}

action conditioning with T5-based text-conditioning during pretraining, denoted as \textbf{T-pre}.
Beyond standard video-generation metrics,
we additionally evaluate action-conditioning fidelity using the optical flow based metric ($\overline{\mathrm{EPE}}$,
$\overline{\cos}$)~\cite{zhang2025reinforcing}, which measures whether the induced motion in generated videos matches the input actions.
Tab.~\ref{tab:wm_eval} reports the results.
Our \arwm{} consistently achieves the best rollout quality across both LIBERO and real-robot data, improving not only appearance metrics but also action-faithfulness.
Together with the qualitative comparison (Fig.~\ref{fig:quali_compare}), these results indicate more accurate dynamics modeling under action-conditioned rollouts.
We also fine-tune \arwm{} for 60k steps and evaluate on RoboNet~\cite{dasari2019robonet}, achieving strong video prediction performance against autoregressive baselines (Tab.~\ref{tab:robonet}).

\noindent \textbf{OOD simulator evaluation}
To further test simulator transfer under distribution shift, we fine-tune \arwm{} on LIBERO and evaluate rollouts on LIBERO-Plus Spatial.
Tab.~\ref{tab:ood_sim_libero_spatial} shows that action-conditioned \bwm{} pretraining improves action-faithfulness metrics over DreamDojo~\cite{gao2026dreamdojo}. The supplement further provides a qualitative OOD comparison, where DreamDojo drifts toward the training-domain appearance while \arwm{} better preserves the novel scene and interaction dynamics.

\begin{table*}[t]
\vspace{-4pt}
\centering
\small
\begin{minipage}[t]{0.45\textwidth}
\centering
\captionsetup{skip=2pt}
\captionof{table}{\textbf{Video prediction evaluation on RoboNet~\cite{dasari2019robonet}.}}
\label{tab:robonet}
\setlength{\tabcolsep}{4pt}
\vspace{-1em}
\resizebox{\linewidth}{!}{%
\begin{tabular}{lcccc}
\toprule
\textbf{Methods} & \textbf{FVD}$\downarrow$ & \textbf{PSNR}$\uparrow$ & \textbf{SSIM}$\uparrow$ & \textbf{LPIPS}$\downarrow$ \\
\midrule
MaskViT~\cite{gupta2022maskvit}   & 211.7 & 20.4 & 67.1 & 17.0 \\
iVideoGPT~\cite{wu2024ivideogpt}  & 197.9 & 23.8 & 80.8 & 14.7 \\
SAMPO~\cite{wang2025sampo}        & \underline{175.3} & \textbf{25.3} & \textbf{84.7} & \underline{12.3} \\
\midrule
\textbf{\arwm{}}                     & \textbf{146.1} & \underline{24.1} & \underline{81.6} & \textbf{8.9} \\
\bottomrule
\end{tabular}}
\end{minipage}\hfill
\begin{minipage}[t]{0.49\textwidth}
\centering
\captionsetup{skip=2pt}
\captionof{table}{\textbf{OOD simulator evaluation on LIBERO-Plus Spatial.}}
\label{tab:ood_sim_libero_spatial}
\setlength{\tabcolsep}{3pt}
\resizebox{\linewidth}{!}{%
\begin{tabular}{lccccc}
\toprule
\textbf{Methods} & \textbf{PSNR}$\uparrow$ & \textbf{SSIM}$\uparrow$ & \textbf{tSSIM}$\uparrow$ & \textbf{EPE}$\downarrow$ & \textbf{cos}$\uparrow$ \\
\midrule
DreamDojo~\cite{gao2026dreamdojo} & 23.79 & .8571 & \textbf{.7778} & .2738 & .2107 \\
\midrule
\textbf{\arwm{}} & \textbf{25.91} & \textbf{.8719} & .7401 & \textbf{.1301} & \textbf{.2761} \\
\bottomrule
\end{tabular}}
\end{minipage}
\vspace{-6pt}
\end{table*}

\begin{wrapfigure}{r}{0.52\columnwidth}
\vspace{-24pt}
\centering
\includegraphics[width=0.42\columnwidth]{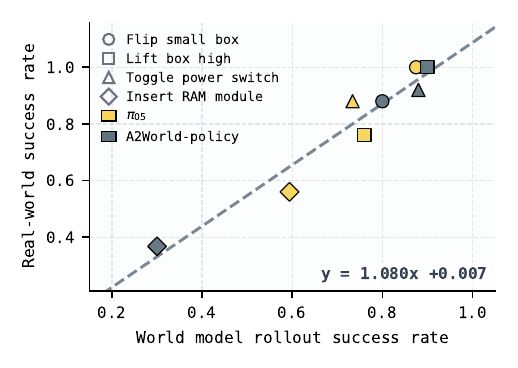}
\vspace{-1.5em}
\caption{\textbf{Simulator consistency on real-robot tasks.} Real-world success rates correlate strongly with \arwm{} rollout success rates across policies and tasks.}
\label{fig:eval_consistency}
\vspace{-20pt}
\end{wrapfigure}

\noindent \textbf{World model as real-world simulator}
We evaluate whether \arwm{} can serve as a real-world simulator for policy evaluation on our robot setup.
Following prior protocols~\cite{guo2025ctrl, gao2026dreamdojo, li2024evaluating, li2025worldeval}, we compare real-world policy success rates with those from closed-loop rollouts in \arwm{} (Fig.~\ref{fig:eval_consistency}).
We run each policy in closed loop inside \arwm{} by autoregressively rolling out observations conditioned on the policy’s action chunks.
Actions are generated at 30 fps and downsampled to 10 fps to match \arwm{}.
Success rates are estimated from $\sim$25 real rollouts and 64 simulator rollouts per policy, with outcomes verified manually.
\arwm{} shows strong agreement with the real world (Spearman $\rho=0.916$, Pearson $r=0.965$, $R^2=0.930$, $N=8$), indicating it is a faithful simulator for policy evaluation.

\subsection{Vision-action joint prediction evaluation}
In this section, we evaluate A2World-policy obtained by directly transferring A2World with only minimal policy-specific changes, without any separate action-only pretraining.
We initialize the policy largely from the pretrained video world model weights, fine-tune it only on downstream robot data, and observe strong performance in both simulation benchmarks and real-robot evaluations.

\noindent \textbf{Evaluation on LIBERO}
We evaluate \vlva{} as a direct policy on LIBERO~\cite{liu2023libero} under the standard 4-suite protocol.
As shown in Tab.~\ref{tab:vlva_libero}, \vlva{} achieves an overall success rate of 98.6\%, outperforming strong baselines with the best average performance.
\begin{table*}[t]
\centering
\begin{minipage}[t]{0.45\textwidth}
\captionsetup{type=table}
\caption{\textbf{Success rate evaluation results on LIBERO~\cite{liu2023libero}.}}
\label{tab:vlva_libero}
\vspace{-0.2em}
\centering
\scriptsize
\setlength{\tabcolsep}{4pt}
\resizebox{\linewidth}{!}{%
\begin{tabular}{lccccc}
\toprule
\textbf{Methods} & \textbf{Spatial} & \textbf{Object} & \textbf{Goal} & \textbf{Long} & \textbf{Average} \\
\midrule
Diffusion Policy~\cite{chi2025diffusion} & 78.3 & 92.5 & 68.3 & 50.5 & 72.4 \\
4D-VLA~\cite{zhang20254d} & 88.9 & 95.2 & 90.9 & 79.1 & 88.6 \\
Dita~\cite{hou2025dita} & 97.4 & 94.8 & 93.2 & 83.6 & 92.3 \\
$\pi_0$~\cite{black2024pi_0} & 96.8 & 98.8 & 95.8 & 85.2 & 94.2 \\
UniVLA~\cite{bu2025univla} & 96.5 & 96.8 & 95.6 & 92.0 & 95.2 \\
$\pi_{0.5}$~\cite{intelligence2025pi_} & \textbf{98.8} & 98.2 & 98.0 & 92.4 & 96.9 \\
OpenVLA-OFT~\cite{kim2025fine} & 97.6 & 98.4 & 97.9 & 94.5 & 97.1 \\
CogVLA~\cite{li2025cogvla} & \underline{98.6} & 98.8 & 96.6 & 95.4 & 97.4 \\
Cosmos Policy~\cite{kim2026cosmos} & 98.1 & \textbf{100.0} & \underline{98.2} & \underline{97.6} & \underline{98.5} \\
\midrule
\textbf{\vlva{}} & 98.2 & \underline{99.2} & \textbf{98.6} & \textbf{98.2} & \textbf{98.6} \\
\bottomrule
\end{tabular}
}
\end{minipage}\hfill
\begin{minipage}[t]{0.49\textwidth}
\captionsetup{skip=3pt}
\caption{\textbf{OOD policy evaluation on LIBERO-Plus Spatial.} }
\label{tab:ood_policy_libero_spatial}
\vspace{-0.2em}
\centering
\scriptsize
\setlength{\tabcolsep}{2pt}
\resizebox{\linewidth}{!}{%
\begin{tabular}{l|ccccccc|c}
\toprule
\textbf{Methods} & \textbf{Bg.} & \textbf{Cam.} & \textbf{Lang.} & \textbf{Light} & \textbf{Layout} & \textbf{Init} & \textbf{Noise} & \textbf{Avg.} \\
\midrule
Cosmos Policy~\cite{kim2026cosmos} & 89.1 & \underline{75.5} & \underline{94.4} & 98.6 & \underline{94.3} & 53.1 & \underline{96.0} & 85.6 \\
FastWAM~\cite{yuan2026fast} & 65.9 & 7.4 & 72.1 & 90.8 & 59.5 & 42.3 & 32.8 & 51.5 \\
$\pi_0$~\cite{black2024pi_0} & \underline{95.0} & 70.7 & 67.9 & 92.8 & 94.0 & 49.1 & 87.7 & 78.6 \\
X-VLA~\cite{zheng2025x} & \textbf{100.0} & 24.5 & 82.1 & \textbf{100.0} & 93.8 & 91.4 & 63.5 & 77.7 \\
GE-Act~\cite{liao2025genie} & 89.1 & 72.3 & 90.8 & \textbf{100.0} & 89.6 & 75.7 & \textbf{97.7} & 87.5 \\
\midrule
\vlva{}-C-init & 83.3 & 53.2 & 90.5 & 96.6 & 89.9 & 77.1 & 74.1 & 80.2 \\
\vlva{}-T-pre & 89.9 & 73.1 & 93.3 & 99.3 & 94.0 & \underline{92.9} & 60.7 & 85.8 \\
\textbf{\vlva{}-A-pre} & 88.0 & 74.2 & \textbf{100.0} & \underline{99.7} & \textbf{94.5} & \textbf{93.1} & 80.9 & \underline{88.5} \\
\vlva{}-P-pre & 89.1 & \textbf{75.8} & \textbf{100.0} & \textbf{100.0} & 92.2 & 91.4 & 72.9 & \textbf{88.6} \\
\bottomrule
\end{tabular}
}
\end{minipage}
\vspace{-0.5em}
\end{table*}

\begin{figure*}
    \includegraphics[width=1.0\linewidth]{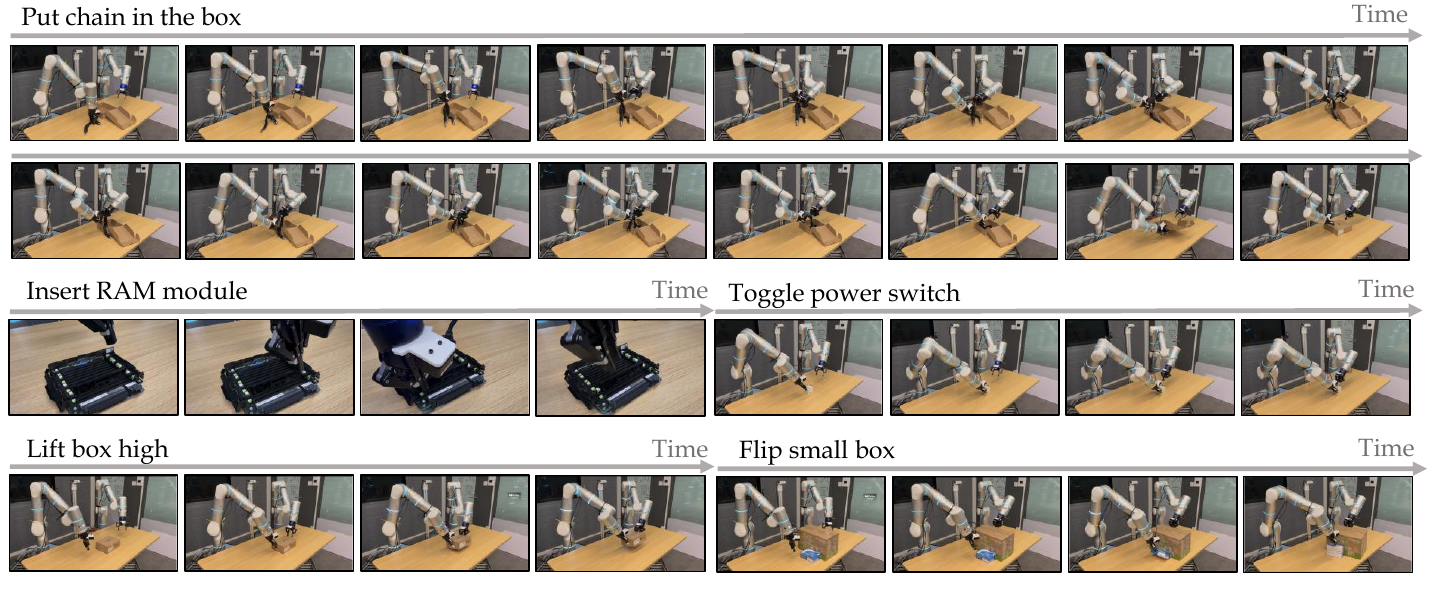}
    \vspace{-2em}
    \caption{\textbf{Real-robot execution results of \vlva{}.} We visualize executions on five real-robot tasks. \vlva{} achieves high success rates across five tasks. In contrast, the baselines~\cite{intelligence2025pi_, lingbot-va2026} often struggle on longer, harder tasks with complex objects (e.g., the chain task), leading to incomplete or unstable executions.}
    \label{fig:real-robot}
\vspace{-0.5em}
\end{figure*}

\begin{figure*}[t]
\centering
    \includegraphics[width=0.95\linewidth]{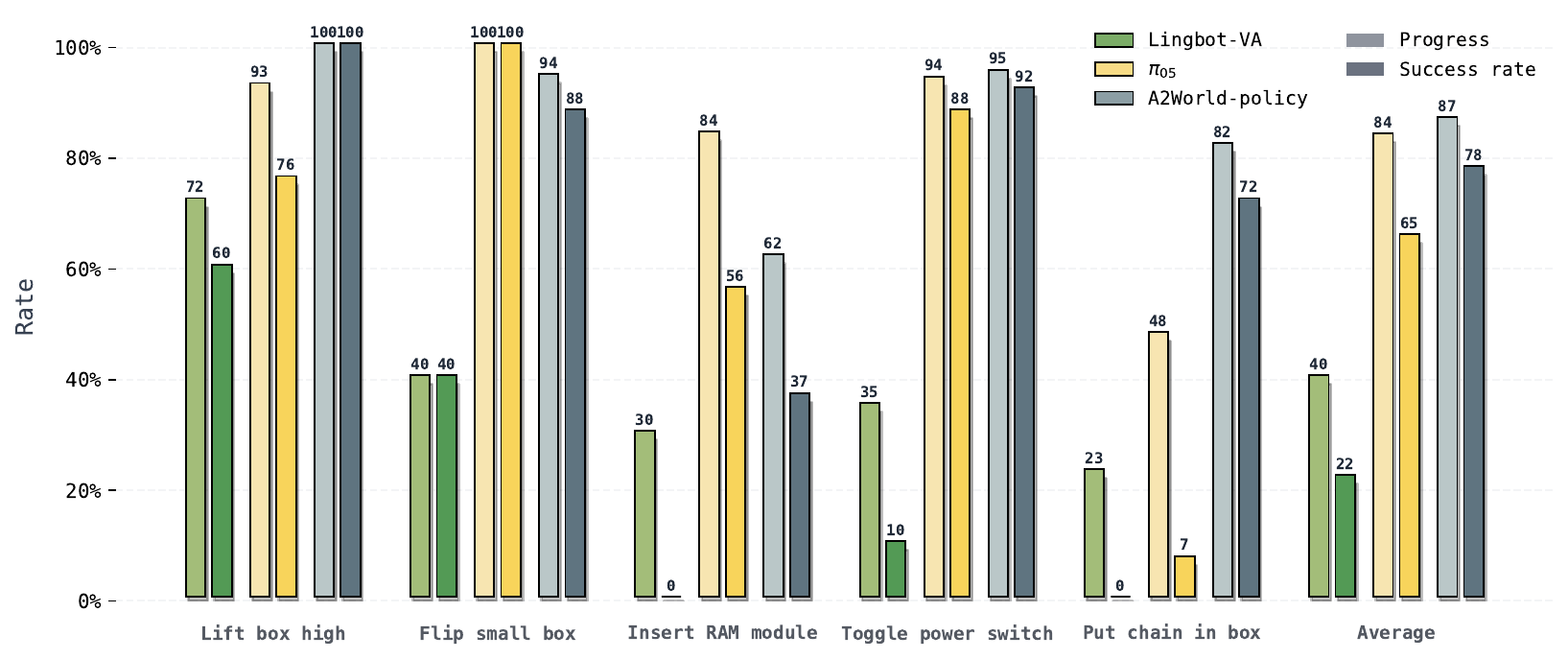}
    \vspace{-1em}
    \caption{\textbf{Detailed real-robot evaluation results.} Our \vlva{} exceeds all baselines in terms of overall task success rate.}
    \label{fig:progress_sr_comparison}
\end{figure*}

\noindent \textbf{OOD policy evaluation}
We evaluate OOD policy transfer by fine-tuning on LIBERO and testing on LIBERO-Plus Spatial~\cite{fei2025libero}, which introduces diverse visual, linguistic, and dynamics shifts.
We compare \textbf{C-init} (directly initializing from Cosmos-Predict2), \textbf{T-pre} (text-conditioned world-model pretraining on the same robot data), \textbf{A-pre} (our action-to-video \bwm{} pretraining), and \textbf{P-pre} (policy-targeted text-video-action pretraining following Eq.~\ref{eq:a2world_policy}).
As shown in Tab.~\ref{tab:ood_policy_libero_spatial}, A-pre reaches 88.5\% average success, clearly outperforming C-init (80.2\%) and T-pre (85.8\%).
It is also essentially on par with P-pre (88.6\%), although P-pre uses a downstream-matched pretraining target.
This indicates that action-to-video pretraining already captures the dynamics prior needed for policy transfer.
More importantly, unlike policy-targeted pretraining, the same A-pre checkpoint is naturally reusable as a long-horizon simulator (Sec.~\ref{sec:arwm-eval}), providing a dual-use prior for both simulator and policy adaptation.

\noindent \textbf{Evaluation on real-robot}
We finally evaluate \vlva{} on a real-robot suite covering diverse contact-rich manipulation, including lifting or reorientation, precision insertion, switch or hinge interactions, and deformable-object handling.
All scores are obtained via third-party evaluation by data-collection operators under a standardized protocol.
Fig.~\ref{fig:progress_sr_comparison} reports task progress and final success, and Fig.~\ref{fig:real-robot} visualizes executions (progress definition in the Appendix).
\vlva{} consistently outperforms strong recent baselines, including $\pi_{0.5}$~\cite{intelligence2025pi_} and LingBot-VA~\cite{lingbot-va2026}, with the largest gains on the most challenging long-horizon, contact-rich tasks where baselines often stall early.

\subsection{Ablations and discussions}
\noindent \textbf{Ablation on history sampling}
We ablate the effect of history sampling on video generation quality by comparing our pose-guided history sampling against:
\textbf{(i)} a standard sliding-window memory strategy that keeps the most recent frames;
\textbf{(ii)} no history injection.
As reported in Tab.~\ref{tab:ab_hist}, pose-guided sampling consistently yields better
rollout quality, indicating that selecting motion-informative histories is important for stable and history-faithful generation.

\begin{table}[t]
\vspace{-6pt}
\centering
\small
\setlength{\tabcolsep}{4pt}

\begin{minipage}[t]{0.48\columnwidth}
\centering
\captionsetup{skip=2pt}
\captionof{table}{\textbf{Results of \vlva{} variants on LIBERO.}}
\vspace{-1em}
\label{tab:vlva_mutual_promotion}
\scalebox{0.65}{
\begin{tabular}{l c}
\toprule
\textbf{Settings} & \textbf{Overall succ.} \\
\midrule
Text-conditioned Cosmos init (C-init) & 97.0 \\
Text-conditioned \bwm{} pretrain (T-pre)  & 97.4 \\
Action-video \bwm{} pretrain (A-pre) & \underline{98.6} \\
\textbf{Policy-targeted pretrain (P-pre)} & \textbf{98.8} \\
\midrule
Video frozen; only shared self-attn trainable & 86.2 \\
\bottomrule
\end{tabular}}
\end{minipage}\hfill
\begin{minipage}[t]{0.49\columnwidth}
\centering
\captionsetup{skip=2pt}
\captionof{table}{\textbf{Ablation on history sampling on LIBERO.}}
\label{tab:ab_hist}
\scalebox{0.58}{
\begin{tabular}{lccccc}
\toprule
\textbf{Methods} & \textbf{PSNR}$\uparrow$ & \textbf{SSIM}$\uparrow$ & \textbf{tSSIM}$\uparrow$ & \textbf{EPE}$\downarrow$ & \textbf{cos}$\uparrow$ \\
\midrule
No history & 25.41 & .8806 & .7663 & .3969 & .5778 \\
Sliding window & 25.63 & .8840 & .7699 & .3900 & .5853 \\
\textbf{Pose-guided (ours)} & 26.64 & .8957 & .7862 & .3498 & .6045 \\
\bottomrule
\end{tabular}}
\end{minipage}

\vspace{-15pt}
\end{table}

\noindent \textbf{Pretraining variants}
Tab.~\ref{tab:vlva_mutual_promotion} compares policy variants after LIBERO fine-tuning.
Policy-targeted pretraining gives the best in-domain average (98.8\%), while our action-video \bwm{} pretraining is very close (98.6\%).
Compared with C-init and T-pre, A-pre benefits from a less ambiguous action-conditioned objective: given the current observation and future actions, the future visual transition is largely determined, whereas text instructions can correspond to many valid action sequences.
This stronger action-to-dynamics prior explains why A-pre transfers better to policy learning than generic video initialization or text-conditioned pretraining.
Together with the OOD results in Tab.~\ref{tab:ood_policy_libero_spatial}, this suggests that a policy-specific pretraining objective mainly provides a small endpoint-specific gain, whereas action-to-video pretraining offers nearly the same policy transfer while also serving as the simulator initialization.

\begin{wrapfigure}{r}{0.48\columnwidth}
\vspace{-22pt}
\centering
\includegraphics[width=0.44\columnwidth]{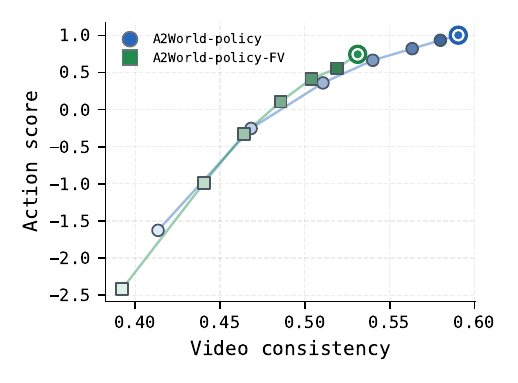}
\vspace{-1.5em}
\caption{\textbf{Video--action coupling during \vlva{} training.} Improving video prediction consistently correlates with better action generation.}
\label{fig:video_action_tradeoff}
\vspace{-16pt}
\end{wrapfigure}

\noindent \textbf{Coupling between video modeling and action learning}
We observe a consistent positive coupling between video prediction and action generation during training.
In Fig.~\ref{fig:video_action_tradeoff}, each point is a validation checkpoint,

where video consistency on future prediction and a normalized action quality score are plotted on the x-axis and y-axis, respectively (definitions in the Appendix).
Better video consistency co-occurs with better action quality, and full joint training reaches a stronger upper-right frontier than freezing the video branch.
Here, the video-frozen variant (86.2\% in Tab.~\ref{tab:vlva_mutual_promotion}) freezes video-specific prediction modules, while keeping the action branch and shared transformer layers trainable.
Overall, the shared representations learned for forecasting and control reinforce each other, improving both video and action.

\section{Conclusions}
We argue that action-conditioned world modeling is a scalable approach for learning transferable dynamics priors.
By leveraging actions as causal supervision, large-scale action-to-video pretraining distills reusable interaction knowledge that generalizes across tasks and environments.
We instantiate this idea with \bwm{}, a multi-view diffusion world model pretrained on diverse real-robot data, and show that its learned prior can be adapted to both \arwm{} for long-horizon rollout and policy evaluation, and \vlva{} for instruction-conditioned control.
Across simulation benchmarks and real-robot experiments, action-to-video pretraining consistently yields a stronger transferable prior than text-conditioned or task-specific robot pretraining.
\section*{Acknowledgements}
This work was supported in part by New Generation Artificial Intelligence-National Science and Technology Major Project (2025ZD0123004), Ningbo grant (2025Z038), and National Natural Science Foundation of China (Grant No.~62376060).

\bibliographystyle{splncs04}
\bibliography{main}

\clearpage
\setcounter{page}{1}
\section{Supplement implementation details}

\subsection{Supplement \bwm{} and \arwm{} details}

\noindent \textbf{Pose-guided history sampling for dual-arm setting}
Alg.~\ref{alg:hist_sampling} in the main text presents the single-arm setting used for LIBERO-style evaluations.
For our dual-arm real-robot setting, we use the variant shown in Alg.~\ref{alg:hist_sampling_bimanual}.
The core idea is unchanged: we compute a pose-induced arc-length along the executed motion and sample frames uniformly in this arc-length space to preserve motion coverage under a fixed history budget.
The only difference is the distance definition: instead of a 6D single-arm pose increment, we use the concatenated dual-arm absolute poses and define step length by summing the weighted translation and rotation changes of both arms, which better reflects coordinated dual-arm motion during sampling.

\noindent \textbf{Detailed parameterization}
For \bwm{}, we use a 35-step sampling schedule with $t_{\min}=0.01$ and $t_{\max}=200.0$. 
We adopt rectified flow with $\sigma_{\min}=4.0$, $\sigma_{\max}=80.0$, and $\rho=7.0$, and set $\sigma_{\text{cond}}$=1e-4, and $\sigma_{\text{data}}=1.0$ (time scaling factor $1.0$), with noise adjustment enabled. 
We scale the final backpropagated loss by $10$.
For multi-view identity, we use learnable view embeddings $\epsilon_{\text{view}}(v)\in\mathbb{R}^{\mathbf{d_e}}$ with $\mathbf{d_e}=7$, concatenated to the latent input channels before patch embedding.
During pretraining, we instantiate 4 view embeddings (up to two third-view and two first-view), matching the maximum camera configuration in our datasets; during fine-tuning, we use the corresponding subset based on available views.
To encourage robustness to view ordering, we randomly shuffle the view order for each training sample.

\subsection{Supplement \vlva{} details}
\vlva{} largely follows the same backbone and video diffusion configuration as \bwm{}, and introduces only a few policy-specific settings.
For our real-robot policy experiments, we use a dual-rate setup with video at 10\,fps and actions at 30\,Hz over a 60-step action horizon.

\noindent \textbf{Additional settings for \vlva{}}
For joint video-action diffusion, we sample a shared base noise level $\sigma_{\text{base}}$ and scale it per modality:
\[
\sigma_v = m_v\,\sigma_{\text{base}}, \qquad
\sigma_a = m_a\,\sigma_{\text{base}},
\]
where $m_v=\sqrt{6}$ and $m_a=0.5$.
We additionally apply a high-$\sigma$ augmentation ratio of $0.05$ on the video branch, while keeping the action noise coupled to the shared $\sigma_{\text{base}}$.
The resulting weighted joint denoising objective is:
\[
\mathcal{L}_{\text{\vlva{}}}
=
\mathbb{E}\!\left[
\mathrm{w}(\sigma_v)\|\hat{\mathbf{z}}_0^v-\mathbf{z}_0^v\|_2^2
+ \lambda_a\,\mathrm{w}(\sigma_a)\|\hat{\mathbf{z}}_0^a-\mathbf{z}_0^a\|_2^2
\right],
\quad
\mathrm{w}(\sigma)=\frac{\sigma^2+\sigma_{\text{data}}^2}{(\sigma\,\sigma_{\text{data}})^2},
\]
with $\lambda_a=1$, and the final backpropagated loss is scaled by $10$.

\begin{algorithm}[t]
\caption{Arc-uniform history sampling (Dual-arm version)}
\label{alg:hist_sampling_bimanual}
\vspace{-0.3em}
\scriptsize
\begin{algorithmic}[1]
\Require History length $T_h$; frame ids $\{f_r\}_{r=1}^{T_h}$; dual-arm absolute poses $\{P_r^{L},P_r^{R}\}_{r=1}^{T_h}$ with $P_r^{(\cdot)}=[p_r^{(\cdot)},\theta_r^{(\cdot)}]\in\mathbb{R}^6$; budget $m$; weights $w_t,w_r>0$; rotation scale $s_r>0$; $\varepsilon>0$
\Ensure Sampled history indices $\mathcal{S}$

\State Set anchor indices $r_s \leftarrow \min\{r \mid f_r=0\}$ (earliest padded/valid frame), $r_e \leftarrow T_h$
\State Define the dual-arm step distance for $r=1,\dots,T_h\!-\!1$:
\[
\Delta p_r^{(\cdot)} \!=\! p_{r+1}^{(\cdot)}-p_r^{(\cdot)},\quad
\Delta \theta_r^{(\cdot)} \!=\! \theta_{r+1}^{(\cdot)}-\theta_r^{(\cdot)},
\]
\[
d_r \leftarrow \Big(
w_t\|\Delta p_r^{L}\|_2^2+w_r\|s_r\Delta\theta_r^{L}\|_2^2
+w_t\|\Delta p_r^{R}\|_2^2+w_r\|s_r\Delta\theta_r^{R}\|_2^2
\Big)^{\frac{1}{2}}.
\]
\State Compute cumulative arc-length: $A_{r_s}\!\leftarrow\!0$; $A_r \leftarrow \sum_{q=r_s}^{r-1} d_q$ for $r=r_s\!+\!1,\dots,r_e$
\If{$|A_{r_e}-A_{r_s}|<\varepsilon$}
    \State \Return $\mathcal{S} \leftarrow$ $\{r_s, r_s,\dots, r_s, r_e\}$ (pad to length $m$)
\EndIf
\State Initialize $\mathcal{S}\leftarrow \{r_s, r_e\}$ and set $n_{\mathrm{mid}}\leftarrow m-2$
\For{$s=1$ to $n_{\mathrm{mid}}$}
    \State $\bar{A}_s \leftarrow A_{r_s} + \frac{s}{m-1}(A_{r_e}-A_{r_s})$
    \State $\hat{r} \leftarrow \arg\min_{r\in(r_s,r_e)} |A_r-\bar{A}_s|$
    \State $\mathcal{S}\leftarrow \mathcal{S}\cup\{\hat{r}\}$
\EndFor
\State \Return chronological indices in $\mathcal{S}$ (sorted; keep anchors)
\end{algorithmic}
\vspace{-0.4em}
\end{algorithm}

\section{Supplement experimental details}
\subsection{Supplement \bwm{} results}
\begin{figure*}
    \includegraphics[width=1.0\linewidth]{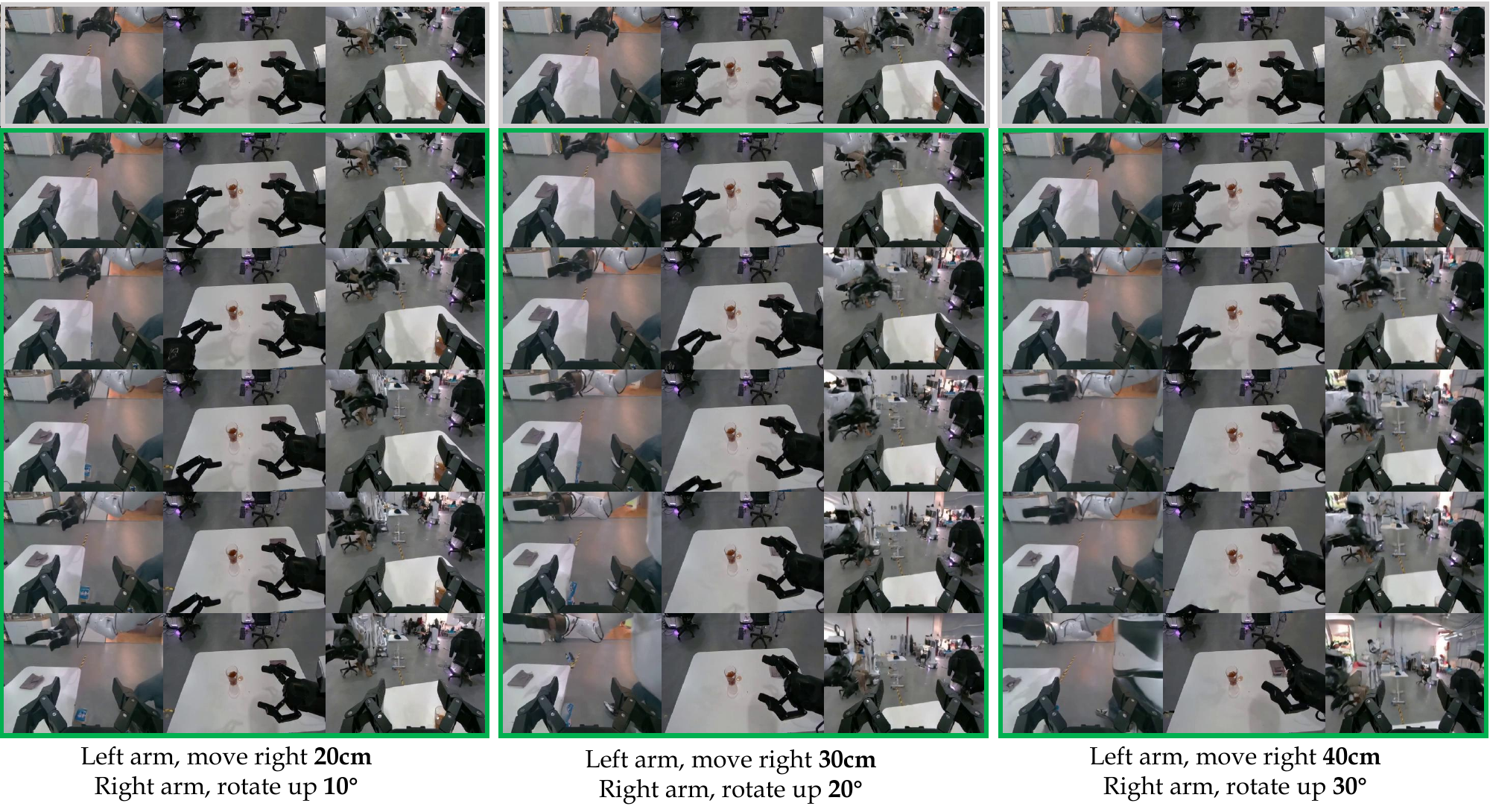}
    \vspace{-2em}
    \caption{\textbf{Precise robot arm control on AgiBot~\cite{bu2025agibot} using \bwm{}.} Starting from the same initial observation, we steer the dual-arm rollout with scripted pose commands of varying magnitudes. \bwm{} follows these continuous controls faithfully and maintains consistent multi-view predictions, demonstrating fine-grained action controllability.}
    \label{fig:agibot_viz}
\vspace{-0.5em}
\end{figure*}

We provide additional qualitative results of \bwm{} in Fig.~\ref{fig:agibot_viz}. 
Starting from the same initial multi-view observation, we apply scripted dual-arm pose commands with increasing magnitudes (e.g., translating the left arm rightward for 20/30/40cm while rotating the right arm upward for 10/20/30°). 
\bwm{} produces rollouts that closely follow these continuous controls and remain coherent across views, demonstrating fine-grained action controllability and stable multi-view consistency beyond task-specific demonstrations.

\subsection{Supplement \arwm{} results}
\begin{figure*}
    \includegraphics[width=1.0\linewidth]{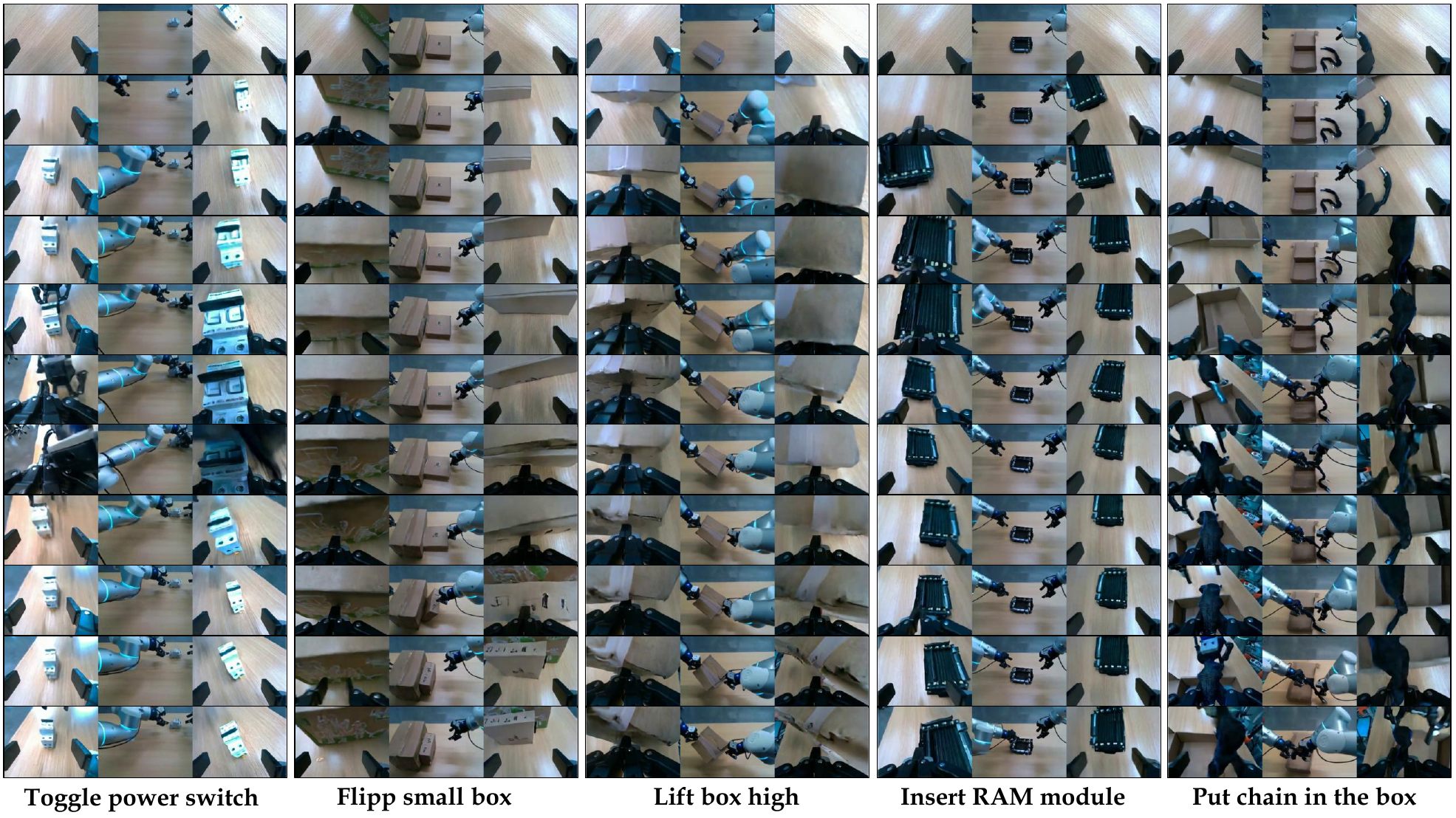}
    \vspace{-2em}
    \caption{\textbf{Closed-loop rollout inside \arwm{} using $\pi_{0.5}$ on our five real-world tasks.} As in Fig.~\ref{fig:quali_compare}, the rollout spans 200 frames (20s at 10fps), and we visualize one frame every 2s.}
    \label{fig:real-world-simulator}
\vspace{-0.5em}
\end{figure*}

\noindent \textbf{World model as real-world simulator}
In Sec.~\ref{sec:arwm-eval}, our real-robot results further indicate that \arwm{} can serve as a reliable real-world simulator for policy evaluation.
Here we additionally visualize long-horizon closed-loop rollouts of $\pi_{0.5}$~\cite{intelligence2025pi_} inside \arwm{}, where actions are inferred step-by-step from the model-generated observations, shown in Fig.~\ref{fig:real-world-simulator}. 
The resulting videos show that \arwm{} preserves spatiotemporal coherence over extended horizons and faithfully reflects success-critical behavior differences (e.g., steady progress versus early stalling or incorrect interactions), supporting simulator-based policy assessment without requiring real-robot execution.

\clearpage

\begin{figure*}[t]
    \centering
    \includegraphics[width=0.6\linewidth]{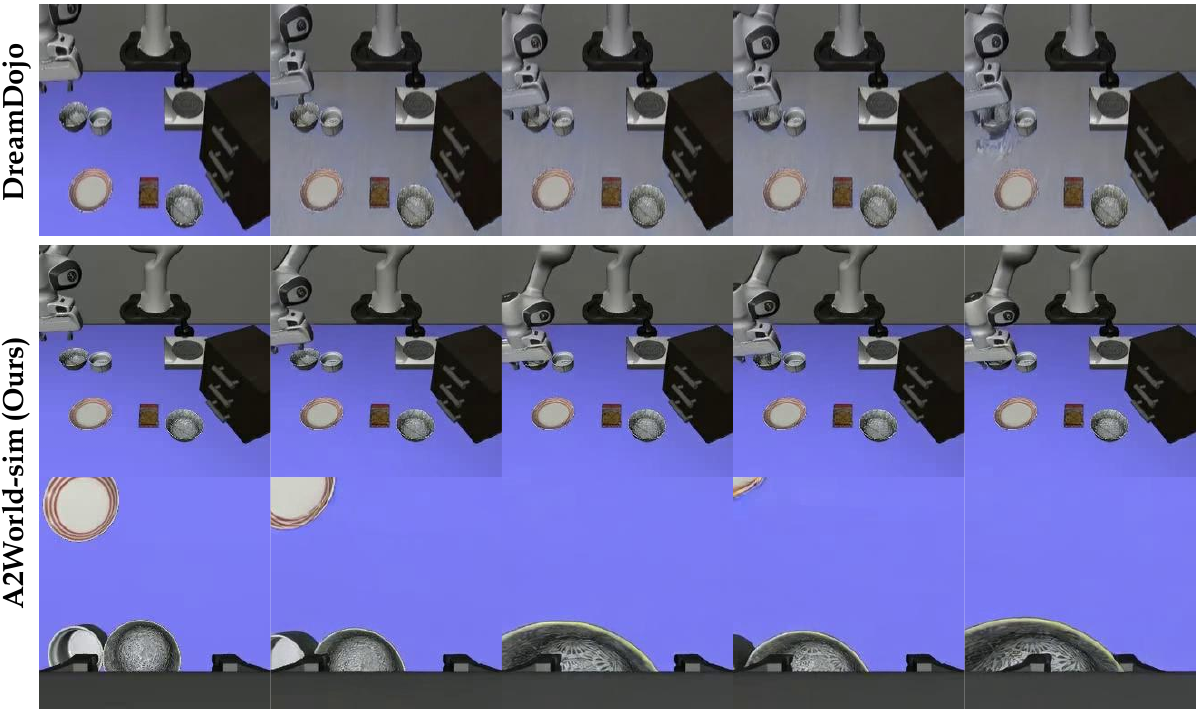}
    \vspace{-0.5em}
    \caption{\textbf{OOD action-conditioned generation comparison.}}
    \label{fig:ood_sim_libero_spatial}
\vspace{-0.5em}
\end{figure*}

\noindent \textbf{Out-of-distribution simulator evaluation}
We further evaluate simulator transfer under an out-of-distribution (OOD) setting by fine-tuning all methods on LIBERO and evaluating rollouts on LIBERO-Plus Spatial.
As shown in Tab.~\ref{tab:ood_sim_libero_spatial} in the main paper, action-conditioned \bwm{} pretraining improves action-faithfulness metrics over DreamDojo~\cite{gao2026dreamdojo}.
Fig.~\ref{fig:ood_sim_libero_spatial} provides a qualitative comparison under an unseen blue-background scene: DreamDojo tends to drift toward the training-domain appearance, whereas \arwm{} better preserves the novel scene and interaction dynamics.

\begin{figure*}
    \centering
    \includegraphics[width=0.9\linewidth]{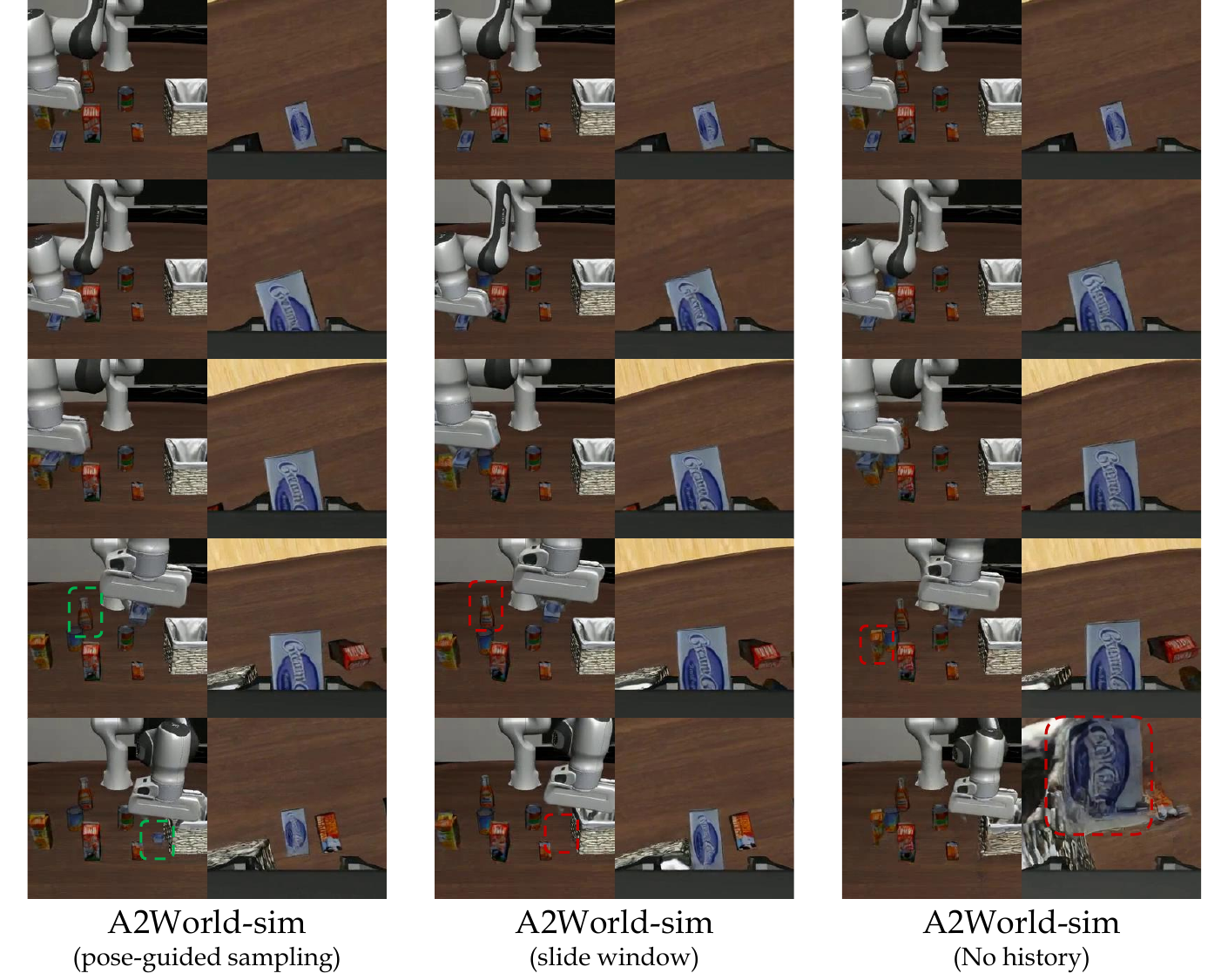}
    \vspace{-1em}
    \caption{\textbf{Qualitative ablation on history sampling.} With a sliding-window history, long rollouts can drift from past object states, causing position shifts or object disappearance; without history this is more severe and may lead to collapse. Pose-guided history sampling largely avoids these artifacts under the same token budget, without introducing extra computational overhead.}
    \label{fig:history_samp}
\vspace{-0.5em}
\end{figure*}

\noindent \textbf{Qualitative ablation on history sampling}
We further visualize the effect of history sampling on LIBERO by comparing our pose-guided history sampling against a standard sliding-window strategy (Fig.~\ref{fig:history_samp}). 

Without injecting motion-informative history, long-horizon rollouts tend to drift and may exhibit failures such as object disappearance or inconsistent object states. 
In contrast, pose-guided sampling selects key motion transitions and interaction states under the same history budget, leading to more stable rollouts with better object permanence and more history-faithful dynamics.

\subsection{Supplement \vlva{} results}

\noindent \textbf{Definition of the axes in Fig.~\ref{fig:video_action_tradeoff}}
Each point in Fig.~\ref{fig:video_action_tradeoff} corresponds to one training checkpoint evaluated on the validation set. 
For each checkpoint, the reported coordinates are obtained by averaging the corresponding metrics over the full validation set.

For the x-axis, let $\hat{V}=\{\hat{I}_t\}_{t=1}^{T}$ and $V=\{I_t\}_{t=1}^{T}$ denote the generated and ground-truth videos on the generated suffix, excluding copied condition-prefix frames. Using Farneb\"ack optical flow, we compute dense flow vectors $\hat{\mathbf{u}}_{t,p},\mathbf{u}_{t,p}\in\mathbb{R}^2$ from frame $t$ to $t+1$ at pixel $p$, and define:
\[
c_{t,p}=\frac{\langle \hat{\mathbf{u}}_{t,p},\mathbf{u}_{t,p}\rangle}
{\|\hat{\mathbf{u}}_{t,p}\|_2\|\mathbf{u}_{t,p}\|_2+\varepsilon},\quad \varepsilon=10^{-6},
\]
together with a validity mask $m_{t,p}\in\{0,1\}$, where $m_{t,p}=1$ if $\|\hat{\mathbf{u}}_{t,p}\|_2>\tau$ and $\|\mathbf{u}_{t,p}\|_2>\tau$ with $\tau$=1e-3, and $m_{t,p}=0$ otherwise. The plotted \emph{Video consistency} is defined as the mean cosine similarity over valid flow locations:
\[
x=
\frac{\sum_{t,p} m_{t,p}\,c_{t,p}}{\sum_{t,p} m_{t,p}}.
\]

For the y-axis, we compute three scalar action-prediction errors from the same checkpoint: translation error $e_{\mathrm{trans}}$, rotation error $e_{\mathrm{rot}}$, and gripper-state error $e_{\mathrm{grip}}$. 
Specifically, $e_{\mathrm{trans}}$ is the mean absolute translation error in meters, $e_{\mathrm{rot}}$ is the mean geodesic rotation error on $\mathrm{SO}(3)$ in degrees, and $e_{\mathrm{grip}}=1-\mathrm{F1}_{\mathrm{grip}}$. 
All action errors are computed after transforming predicted and target actions back to physical units via the inverse of dataset normalization. 
These three errors are normalized by z-score over all checkpoints plotted in Fig.~\ref{fig:video_action_tradeoff}, pooling all compared methods:
\[
z(q)=\frac{q-\mu_q}{\sigma_q},\qquad q\in\{e_{\mathrm{trans}},e_{\mathrm{rot}},e_{\mathrm{grip}}\}.
\]
The plotted action quality score is:
\[
y
=-\frac{z(e_{\mathrm{trans}})+z(e_{\mathrm{rot}})+z(e_{\mathrm{grip}})}{3}.
\]
Thus, larger $y$ indicates better overall action-prediction quality under this combined metric. 
Because the normalization is performed over the checkpoints shown in this figure, this score is intended for relative trend comparison within Fig.~\ref{fig:video_action_tradeoff}, rather than absolute comparison across different figures or experiments.

\begin{figure*}
    \includegraphics[width=1.0\linewidth]{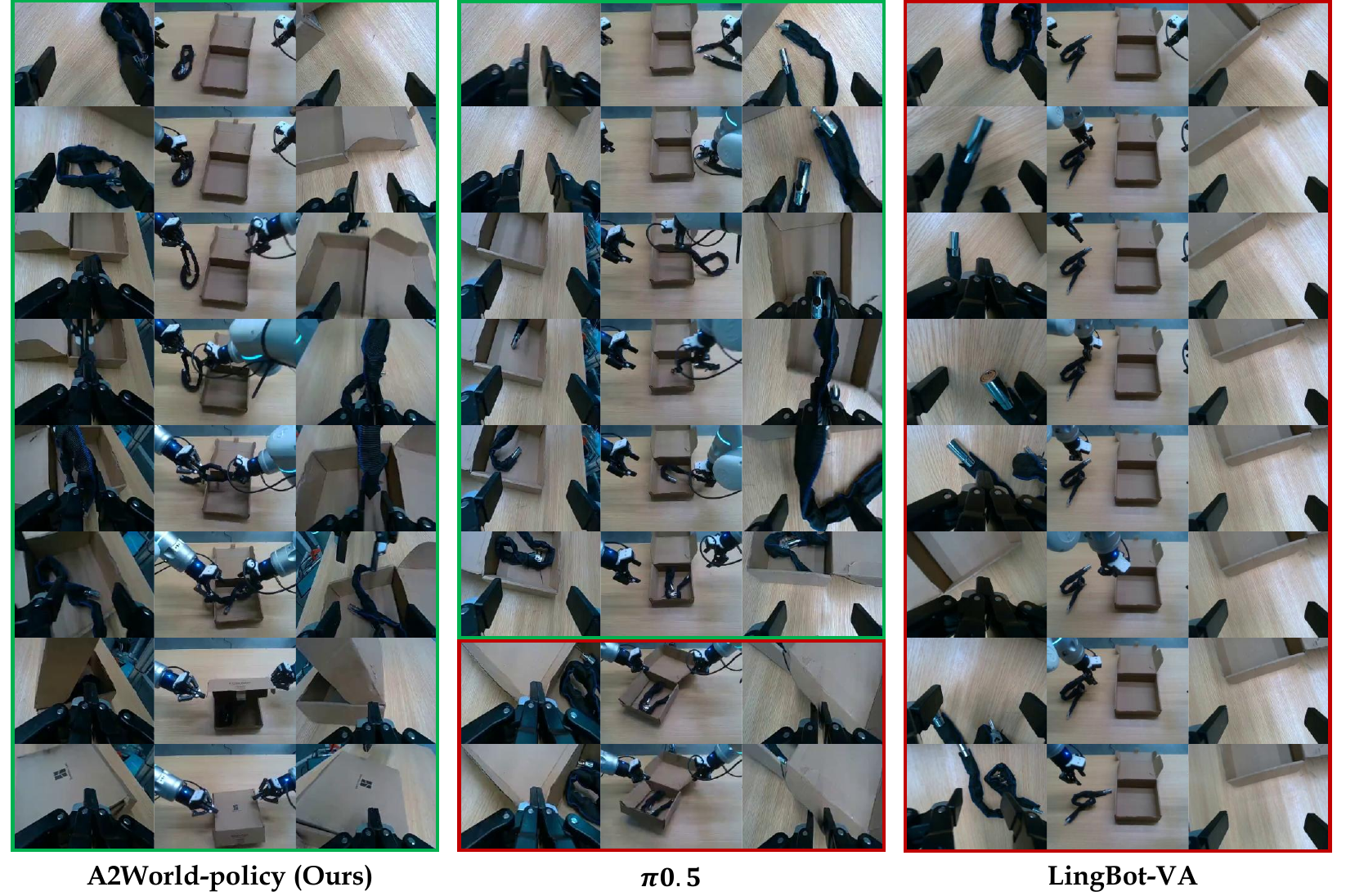}
    \vspace{-2em}
    \caption{\textbf{Supplementary real-robot execution on \emph{Put chain in the box}.} Qualitative comparisons between \vlva{} and two baselines ($\pi_{0.5}$~\cite{intelligence2025pi_} and LingBot-VA~\cite{lingbot-va2026}). \vlva{} more consistently completes the full sequence, while baselines often fail to place the chain or cannot finish closing the box.}
    \label{fig:real-world-vlva}
\vspace{-0.5em}
\end{figure*}

\noindent \textbf{Qualitative real-robot results}
We provide additional qualitative real-robot results in Fig.~\ref{fig:real-world-vlva}, focusing on the long-horizon and contact-rich task \textit{put chain in the box}. 
Together with the quantitative progress and success statistics in Fig.~\ref{fig:progress_sr_comparison}, the qualitative rollouts highlight that \vlva{} executes the multi-stage interaction more reliably and consistently.
In particular, \vlva{} is able to complete the full task sequence, including both inserting the deformable chain and closing the box, whereas the baselines~\cite{intelligence2025pi_, lingbot-va2026} often fail in two typical ways: they either reach a partial completion state (e.g., the chain is placed inside but the box is not closed), or fail at the fundamental step (e.g., cannot place the chain into the box at all).
These results suggest that \vlva{} is more robust to long-horizon error accumulation and better handles precise state transitions under rich contacts, which is crucial for challenging real-world manipulation.

\noindent \textbf{Task progress definition}
For Fig.~\ref{fig:progress_sr_comparison}, we assign each rollout a normalized progress score in $[0,1]$ by manually mapping its final state to task-specific completion milestones, where $1.0$ denotes full success. 
\textbf{(i)} For \textit{Lift box high}, lifting the box successfully is scored as $1.0$, contact without lifting is scored as $0.3$, partial but incomplete lifting is scored between $0.5$ and $0.7$ depending on height and stability. 
\textbf{(ii)}  For \textit{Flip small box}, a successful flip is scored as $1.0$, cases where the box is only squeezed or pushed upward without being flipped are scored as $0.4$. 
\textbf{(iii)}  For \textit{Insert RAM module}, full insertion on both sides is scored as $1.0$, half insertion is scored as $0.5$, partial insertion with a clear pressing trend is scored as $0.6$ or $0.7$ depending on insertion depth. 
\textbf{(iv)} For \textit{Toggle power switch}, fully toggling the switch to the target state is scored as $1.0$, lifting the switch without completing the toggle is scored as $0.4$, near-complete toggles that stop short of the target state are scored as $0.7$. 
\textbf{(v)}  For \textit{Put chain in box}, we use finer-grained stages due to the longer horizon and richer contacts: no contact with an approaching tendency is scored as $0$, initial contact as $0.1$ or $0.2$, lifting the chain as $0.3$, partial insertion into the box as $0.4$, full insertion as $0.5$, contacting the lid as $0.6$ or $0.7$, partial closing as $0.8$, and full completion as $1.0$.

\end{document}